%% file: paper.tex
\documentclass[journal]{IEEEtran}  

\usepackage{graphicx} 
\usepackage{amsmath} 
\usepackage{amssymb}  
\usepackage{algorithm}
\usepackage[T1]{fontenc}
\usepackage{algpseudocode}
\usepackage{xcolor}
\usepackage{nicefrac}
\input{packages}
\newtheorem{thm}{Theorem}[section]

\newcommand{\qed}{$\blacksquare$}

\newcommand{\bydef}{\triangleq}
\newcommand{\probd}{\mathbb{P}}
\newcommand{\prob}{\mathrm{P}} 
\newcommand{\obj}{\mathcal{U}}

\DeclareMathOperator*{\argmax}{arg\,max}
\definecolor{ultraviolet}{RGB}{95, 75, 139}
\definecolor{deepteal}{RGB}{0, 51, 51}
\definecolor{hibiscus}{RGB}{176,48,96}
\definecolor{emeraldgreen}{RGB}{0,155,119}
\definecolor{violet}{RGB}{238,130,238}
\definecolor{inkscapePurple}{RGB}{128,0,128}
\definecolor{inkscapeOlive}{RGB}{128,128,0}
\definecolor{indigo}{rgb}{75,0,130}

\title{\LARGE \bf
Simplified Continuous High Dimensional Belief Space Planning with Adaptive Probabilistic Belief-dependent Constraints
}

\author{Andrey Zhitnikov$^{1}$ and Vadim Indelman$^{2}$ \\
\thanks{This work was partially supported by the Israel Science Foundation (ISF).}
$^1$Technion Autonomous Systems Program (TASP) \\
$^2$Department of Aerospace Engineering\\
Technion - Israel Institute of Technology, Haifa 32000, Israel\\
\small{\texttt{andreyz@campus.technion.ac.il, vadim.indelman@technion.ac.il}}}

\begin{document}

\maketitle
\thispagestyle{empty}
\pagestyle{empty}

\begin{abstract}
Online decision making under uncertainty in partially observable domains, also known as Belief Space Planning, is a fundamental problem in robotics and Artificial Intelligence. Due to an abundance of plausible future unravelings, calculating an optimal course of action inflicts an enormous computational burden on the agent. Moreover, in many scenarios, e.g., information gathering, it is required to introduce a belief-dependent constraint. Prompted by this demand, in this paper, we consider a recently introduced probabilistic belief-dependent constrained POMDP. We present a technique to adaptively accept or discard a candidate action sequence with respect to a probabilistic belief-dependent constraint, before expanding a complete set of future observations samples and without any loss in accuracy. Moreover, using our proposed framework, we contribute an adaptive method to find a maximal feasible return (e.g., information gain) in terms of Value at Risk for the candidate action sequence with substantial acceleration. On top of that, we introduce an \emph{adaptive simplification} technique for a probabilistically constrained setting. 
Such an approach provably returns an identical-quality solution while dramatically accelerating online decision making.   
Our universal framework applies to any belief-dependent constrained continuous POMDP with parametric beliefs, as well as nonparametric beliefs represented by particles. In the context of an information-theoretic constraint, our presented framework stochastically quantifies if a cumulative information gain along the planning horizon is sufficiently significant (e.g.~for, information gathering, active SLAM). We apply our method to active SLAM, a highly challenging problem of high dimensional  Belief Space Planning. Extensive realistic simulations corroborate the superiority of our proposed ideas.
\end{abstract}

\section{Introduction}

A comprehensive approach to craft many online decision-making problems, characterized by the agent situated in an environment and acting under uncertainty, is the Partially Observable Markov Decision Process (POMDP). For most such problems, it is sufficient to assume that the belief-dependent reward is merely the expectation of a state-dependent reward with respect to belief. This assumption is the case in classical POMDP formulations. In contrast, numerous problems in robotics, such as informative planning tasks \cite{Hollinger14ijrr}, active Simultaneous Localization and Mapping (SLAM) \cite{Placed22arxiv2},  and sensor placement problem \cite{Kopitkov17ijrr} 
are explicitly concerned with decreasing uncertainty, thereby raising the need for planning with general belief-dependent reward functionals. 

General belief-dependent operators were examined in the context of reward but hardly so in the context of the constraint. 
In the robotics community, continuous POMDP with belief-dependent information theoretic rewards is known as Belief Space Planning  (BSP) \cite{VanDenBerg12ijrr, Indelman15ijrr}. One of the embodiments of BSP, and also the subject of our interest, is active SLAM. In this context, the robot's state comprises the robot's pose trajectory and the map to be estimated. In the Gaussian parametric full SLAM problem, the previous robot poses are not marginalized out but kept to preserve the sparsity of the information matrix \cite{Dellaert17foundations}.

Since belief is to be maintained over an increasingly high-dimensional state, it is not an easy task for an online operating robot. Keeping the whole robot's trajectory causes the state dimension to grow with time. Moreover, the fact that the agent's environment is part of the state, and the agent reveals more area with operation time, also contributes to the state dimensionality. This computational challenge is known as \emph{curse of dimensionality}. 
With an increasing planning horizon, the number of possible measurements and candidate action sequences grows exponentially, assembling the computationally intractable decision making problem. This phenomenon is usually regarded as the \emph{curse of history}. Many research efforts have targeted both \emph{curses}.

The abundance of possible future observations within the planning phase is often resolved, in robotics, by the Maximum Likelihood (ML) assumption \cite{Platt10rss}. While widely used, taking into account merely the most likely measurements is highly unrealistic, particularly in the presence of significant uncertainty.  It is possible that the largest available reward is not the most likely one, resulting in a substantial error in the objective estimate and consequently suboptimal autonomous behavior. One standing out approach to use a number of sampled observations instead of ML assumption alongside generic belief-dependent rewards builds upon reuse of calculations, alleviating the computational burden \cite{Farhi21arxiv}, \cite{Farhi19icra}.

The Artificial Intelligence community also engaged in augmenting the  classical POMDP formulation with belief-dependent rewards. The journey started from $\rho$-POMDP \cite{AryaLopez10nips} and significantly advanced through time \cite{Fehr18nips}, \cite{Dressel17icaps}, \cite{Sunberg18icaps}.

Recent methods, merging  both worlds, build upon the \emph{simplification} paradigm \cite{Sztyglic21arxiv_b}, \cite{Zhitnikov22ai}, \cite{Sztyglic22iros}.  These simplification-based methods finally relax limiting assumptions, e.g., Gaussian belief, Piecewise linearity, or Lipshitz continuity of the reward,  and permitted universal belief-dependent rewards such as differential entropy of general beliefs. Since the differential entropy operator acts over the belief, which can be parametrized in various ways, e.g., Gaussian or set of particles, questions of Piecewise linearity or Lipshitz continuity are vague and well defined only when the state is discrete and finite. In a continuous setting they shall be approached individually for each belief parametrization. This fact discards many early approaches \cite{AryaLopez10nips},  \cite{Fehr18nips} to include belief-dependent rewards to POMDP.   
Another line of \emph{simplification} works alleviate curse of dimensionality in the setting of multivariate Gaussian distributions utilizing sparsification \cite{Elimelech22ijrr} and topological \cite{Kitanov19arxiv} aspects. The simplification paradigm was also applied with Gaussian-mixture distributed beliefs \cite{Shienman22icra}, \cite{Shienman22isrr}. Another important mechanism for the interplay between computational effort and the quality of the result is adaptivity \cite{Barenboim22ijcai}.

All discussed above decision-making methods are concerned with selecting the optimal action, disregarding the actual amount of profit or risk entirely. However, it is  essential since preventing the robot from performing  unnecessary or self-destructive operations is highly important. This gap can be filled by introducing constraints into decision-making formulation. Some attempts to do so in the context of safe POMDPs include chance constraints \cite{Santana16aaai}.  

A general belief-dependent constraint, however, has not received proper attention so far except in our previous work \cite{Zhitnikov22arxiv}, where we focused on safety and not information gathering tasks. 

In this paper, we continue to investigate the facets of our proposed earlier framework \cite{Zhitnikov22arxiv} of general belief-dependent constrained continuous POMDP. Motivated by information gathering, also called informative planning tasks, we focus on the cumulative form of the constraint in contrast to the multiplicative form as in our previous paper.   
One of the specific applications of our framework is stopping exploration. Moreover we provably extend the simplification framework to both forms of the constraints in our novel probabilistically constrained setting.  The first form is cumulative and the second is multiplicative.
   
There are attempts to use differential entropy gain as a constraint to halt exploration in the problem of active SLAM \cite{Cadena16arxiv} \cite{Placed22arxiv}. However, it was never fully explored since, typically, algorithms solving BSP, in particular, active SLAM under partial observability, assume Maximum Likelihood observations  \cite{Placed22arxiv} to alleviate the computational burden.  Stopping exploration is still regarded as an open problem \cite{Cadena16arxiv}.

Our probabilistic belief-dependent constraint of cumulative form, which will become apparent later,  generalizes previous approaches. The naive way to threshold a belief-dependent operator under partial observability is to perform expectation with respect to observations. However, even this has gained less attention so far and has not been done to the best of our knowledge since commonly existing approaches take into account only ML observations. In contrast to expectation with respect to future observations, we propose a probabilistic constraint. Our proposed variant is sensitive to the distribution of the belief-dependent constraint, while averaging with respect to future observations is not. 

As opposed to a threshold on expectation with respect to observations, we propose two conditions. Interior condition thresholds using $\delta$ the belief-dependent operator (return) for given sequence of possible future observations. The exterior condition verifies that the interior one is satisfied with  confidence level of at least $1-\epsilon$. To rephrase it, we require that the fraction of the observation sequences fullfilling the interior condition will be at least $1-\epsilon$.   In due course, we consider two different problem formulations. In the first problem, $\delta$ is specified externally by the user. We coin this problem as {\bf optimality under a probabilistic constraint}. In the second problem, that we name {\bf maximal feasible return},  $\delta$  is a free parameter. In turn, our formulation and approach enable fast adaptive maximization of Value at Risk (VaR) on top of a general belief-dependent return. This problem is highly challenging due to the fact that VaR is not a coherent functional \cite{Pflug16MOR}.

Our contributions are fourfold. First, we utilize our probabilistically constrained Partially Observable Markov Decision Process (POMDP) in the context of information-theoretic constraint. We analyze Mutual Information constraint in this context, following the expectation approach versus our novel probabilistic constraining. Notably, we did not find any works shifting the Mutual Information from the reward operator to the constraint. Second, we cast unconstrained Risk Aware Belief Space Planning with Value at Risk as a purely constraints-driven problem. As we unveil in this paper, this reformulation enables the decision maker to save time by adaptively expanding the lowest required number of observations without compromising the quality of the solution. Third, we rigorously derive a theory of the \emph{simplification}. Given a converging to the constraint and reward bounds, our approach can be simplified, gaining substantial speedup without any loss in performance. We apply our technique to High Dimensional Belief Space Planning. In particular, our case study is active SLAM.
%

The remainder of this paper is structured as follows. We start from background and notations in section \ref{sec:BackgroundNotations}. Our next step is the in depth discussion of the problem formulation and our approach (section \ref{sec:Approach}). We then present and application of our methods (section \ref{sec:Application}) and continue to the simulations and results (section \ref{sec:Sim}). The conclusions section \ref{sec:Conclusions} finalizes the paper   We placed the proofs in the appendix to preserve continuity flowing and remove clutter from the central paper ideas.  

\section{Background and Notations} \label{sec:BackgroundNotations}
Let us formally introduce the Partially Observable Markov Decision Process with belief-dependent rewards named $\rho$-POMDP  alias to BSP. By the bold symbols, we denote time vector quantities; by $\square_{a:b}$, we mark series annotated by the time discrete indices running from $a$ to $b$ inclusive.  By $\probd$ we denote probability density function and by $\prob$ the probability. By lowercase letter we denote the random quantities or the realizations depending on the context.  
  
The $\rho$-POMDP is a tuple $(\mathcal{X}, \mathcal{A}, \mathcal{Z}, T, O, \rho, \gamma, b_0)$ where $\mathcal{X}, \mathcal{A}, \mathcal{Z}$ denote state, action,
and observation spaces with $x \in \mathcal{X}$, $a \in\mathcal{A}$, $z \in \mathcal{Z}$ the momentary state, action, and observation, respectively. $T(x', a, x) = \probd_T(x' | x, a)$ is a stochastic transition model from the past state $x$ to the subsequent $x'$ through action $a$.  
So far, we have described the classical components of POMDP. However, in BSP, the observation model $O(\cdot)$ undergoes a customization that will be apparent later. For now, we left it undefined.   
Further, $\gamma \in (0, 1]$ is the discount factor, $b_0$ is the belief over the
initial state (prior), and $\rho$ is the belief-dependent reward operator. 

For conciseness let us denote interchangeably $\square_{k+}$ and $\square_{k:k+L-1}$, as well as $\square_{(k+1)+}$ and $\square_{k+1:k+L}$.  In this paper, we deal with static action sequences of variable horizon $L$. Namely, our action space is 
%
$	\mathcal{A} \bydef \{a^i_{k:k+L^i-1}\}^{|\mathcal{A}|}_{i=1}$.
%
Our actions along a particular action sequence are of different lengths. We also can think about such an action sequence as a path $\mathcal{P}$ comprising motion primitives. However, the action sequence is a much more general notion.   

An autonomous robot deployed in an unknown environment repeatedly performs acting, sensing, and planning sessions up until it reaches the required goal or \emph{fails} to do so as we further formulate. 

Let $h_{t}$ denote history of actions $a_{0:t-1}$ and observations $z_{1:t}$ obtained by the agent up to time instant $t$ and the prior belief $b_0$. To clarify, we denote by $t$ an arbitrary time instant and by $k$ the time instant of the current planning session. Such that if $t\geq k$, the subscript $t$ regards to future time. Another representation of history is the posterior belief. We define the posterior belief $b_{t}$ as a shorthand for  
the probability density function of the state $\boldsymbol{x}_{t}$ given all information up to time instant $t$, i.e.~
%
	$b_{t}(\boldsymbol{x}_{t}) \bydef \probd(\boldsymbol{x}_{t}|h_{t})$.	
%
In this paper the belief converts the history to a more convenient form, $b_t$ and can be used interchangeably with $h_t$, as opposed to our previous work \cite{Zhitnikov22ai}. 

Often times in BSP problems, the robot's map is unknown and therefore regarded as a random quantity. This allows the robot to operate in unfamiliar environments. We opt for landmarks map representation, so the robot's state is 
\begin{align}
	\boldsymbol{x}_{t} \triangleq \Big(x_{0:t}, \{\ell^j\}_{j=1}^{M(k)}\Big), \label{eq:State}
\end{align}
where $M(k)$ is the number of landmarks the robot has observed until time instant $k$ inclusive. These landmarks represent the unknown robot's environment, specifically the map, to be estimated.  
To emphasize that $j$ is not a time index, we denote it by a superscript instead of a subscript. 
\begin{figure}[t] 
	\centering
	\begin{minipage}[t]{0.49\columnwidth}
		\centering
		\includegraphics[width=\columnwidth]{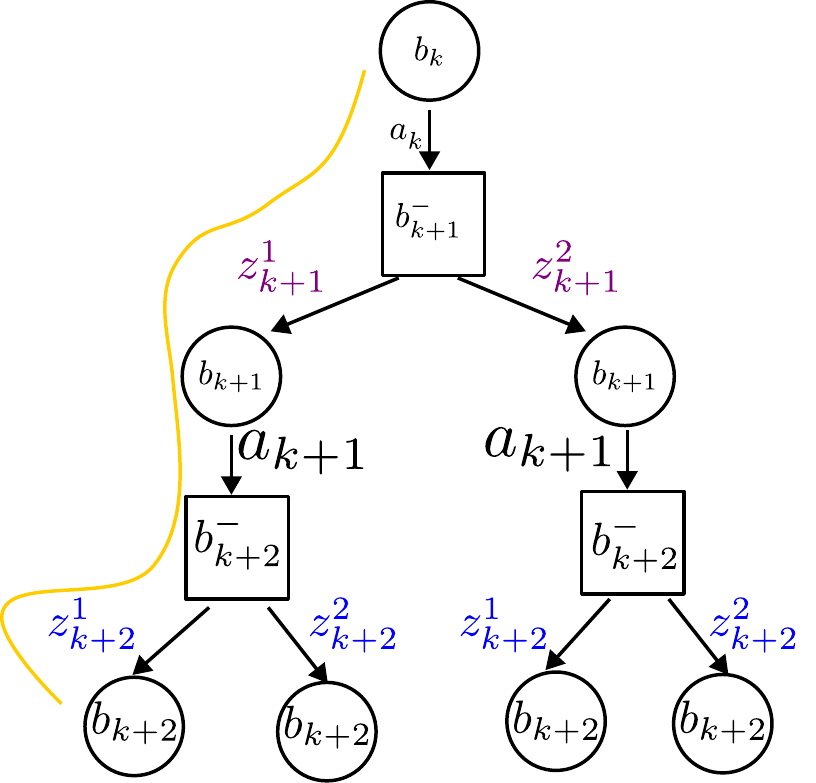}
		\subcaption{}
		\label{fig:BeliefTreeBeta}
	\end{minipage}%
	\hfill
	\begin{minipage}[t]{0.49\columnwidth}
		\centering 
		\includegraphics[width=\textwidth]{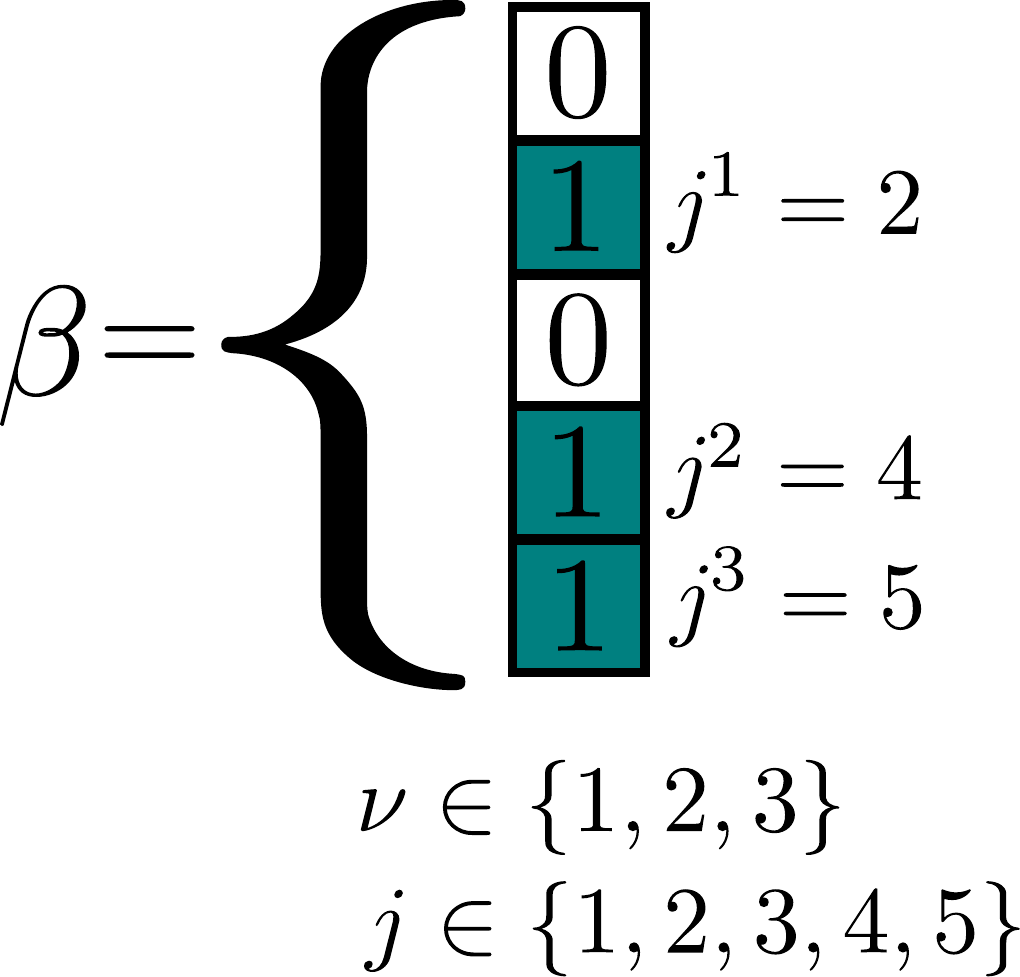}
		\subcaption{}
		\label{fig:Beta}
	\end{minipage} 
	\caption{ \textbf{(a)} Visualization of the belief tree given the realization of $\beta_{k+1:k+2}$ for action sequence $a_{k+}$. Here we show two samples of observations per propagated belief. By {\color{inkscapePurple}{\bf{purple}}} and {\color{blue}{\bf{blue}}} colors, we denote possibly different dimensionality of the observation stemming from the number of visible landmarks. The yellow lace illustrate the observation sequence $z_{k+1:k+2}$ (Section \ref{sec:AdaptiveBeliefTree}); \textbf{(b)} One possible realization of configuration is $\beta=(01011)^T$.  } 
\end{figure}
\subsection{Random landmarks configuration vector}
Let  $\beta_t \in \{0,1\}^{M(k)}$ be a random vector of  Bernoulli variables, statistically independent given robot's pose $x_{t}$, as will be shortly displayed by \eqref{eq:ConfigModel} and \eqref{eq:TotalConfigModel}. Its dimensionality is the number of landmarks present in the belief. Each realization of $\beta_t$ defines a subset of visible landmarks. Such a realization has ones at the indexes of visible landmarks and zeros else, such that $[\beta]^{j} = 1, \forall j \in \{j^{\nu}\}_{\nu=1}^{n(\beta)}$, where $n(\beta)= \sum_{j} [\beta]^j $. (By $[\cdot]^{j}$ we indicate the coordinate $j$ of a vector.) The superscript $\nu$ defines a subsequence of indices $j^{\nu}$ of visible landmarks (Fig.~\ref{fig:Beta}). Let us clarify, $j^1, j^2,\dots$ represent the  strictly increasing with $\nu$ values of indexes of enumerated landmarks resulting in a random set $\{j^{\nu}\}_{\nu=1}^{n(\beta)}$, such that $j^{\nu} = j(\nu)$. 

The mapping from the Boolean vector $\beta$ to the random finite set of indices $\{j^{1}, j^2 \dots\}$ is invertible.  Therefore, one can define a probability over the random finite sets  \cite{Mullane11tro} instead of boolean vectors. 

One way to define a probabilistic model for visible landmarks configuration given $x$ is  
\begin{equation}
	\label{eq:ConfigModel}
	\begin{split}
		&\prob_{\beta}([\beta_{t}]^{j}=1|x_{t}, \ell^{j}) =\mathbf{1}_{ \{\|x_{t} - \ell^{j}\| \leq r\}}, \\
		&\prob_{\beta}([\beta_{t}]^{j}=0|x_{t}, \ell^{j}) =1-\mathbf{1}_{ \{\|x_{t} - \ell^{j}\| \leq r\}},
	\end{split}
\end{equation}
where  $r$ is a visibility radius. Our approach is not limited to this specific model and supports any other model; for instance in more complex scenarios Eqs.~\eqref{eq:ConfigModel} would imitate a camera field of view.
Eqs.~\eqref{eq:ConfigModel} portray that each landmark deterministically has a visibility radius.  If the robot is close enough, it receives the signal from the landmark.  Overall we arrive at 
\begin{align}
	\prob_{\beta}\bigg(\beta_{t}\bigg|x_{t},\{\ell^j\}_{j=1}^{M(k)}\bigg) =\prod_{j=1}^{M(k)}\prob_{\beta}([\beta_{t}]^{j}|x_{t}, \ell^j). \label{eq:TotalConfigModel}
\end{align}
Here, we assumed that $t \geq k$ and the planner does not reveal  new landmarks in planning session, that is, $M(k)$ depends on the present time but not the future time $t$.

Now we are ready to define a customized  observation model. 
This model is used in planning session
\begin{align}
	O(z , \boldsymbol{x}, \beta) \bydef \probd(z | \boldsymbol{x}, \beta) = \prod_{\nu=1}^{n(\beta)} \probd_Z(z^{\nu}|x, \ell^{j^{\nu}}). \label{eq:TotalObsModel}
\end{align}
%
\subsection{The objective}
A common BSP objective is given by 
\begin{align}
	&\obj(b_k, a_{k+}) = \underset{\beta_{(k+1)+}}{\mathbb{E}} \!\Bigg[ \obj^{\beta_{(k+1)+}}(b_k, a_{k+})  \Bigg| b_k, a_{k+} \Bigg], \label{eq:Utility}
\end{align}
where 
\begin{align}	
	&\!\!\obj^{\beta_{(k+1)+}}(b_k, a_{k+}) = \label{eq:UtilityBeta} \\
	&\underset{z_{(k+1)+}}{\mathbb{E}} \Bigg[\!\sum_{t=k}^{k+L-1}\!\!\! \! \rho(b_{t},a_t, {z}_{t+1},  b_{t+1})\!\Bigg| b_k,a_{k+}, \beta_{(k+1)+} \Bigg],	\nonumber
\end{align}
and where $t$ is the running time index and  $k$ is the present time instant. 
The inner expectation $\obj^{\beta_{(k+1)+}}(b_k, a_{k+})$ (Fig.~\ref{fig:BeliefTreeBeta}) corresponds to the utility conditioned on static set of visible landmarks. Therefore, per time index, the dimension of the observation is static (It can be different, however, per time index). Thus, the expectation operator is well defined. The outer expectation performs, weighted in terms of $\beta_{(k+1)+}$ (Fig.~\ref{fig:Beta}) average of such values. Note that, while it is appealing to fold the conditional expectations in \eqref{eq:UtilityBeta} using the law of total expectation, we cannot do that since the dimension of the observation $z_t$ depends on the number of visible landmarks represented by each specific realization of $\beta_t$.
By $\rho(b, a, z', b')$ we denote a general-belief dependent reward depending on two consecutive beliefs and the elements relating them.  Further, for clarity, we omit the dependency on the action and the observation.

To summarize this section,  BSP accommodates varying dimension of observation conditioned on state and continuous spaces. 

\section{Problem Formulation and Approach} \label{sec:Approach}
In this work we tackle two problems.   
Our {\bf first} problem is the  {\bf optimality under a probabilistic constraint}   
\begin{equation}
\begin{split}
&   a^* \in \arg\max_{a_{k+} \in \mathcal{A}} \  \obj(b_k, a_{k+}) \text{ subject to}  \\
& \prob (c(b_{k:k+L}; \phi, \delta)=1 | b_k, a_{k+}) \geq 1-\epsilon ,
\end{split}
\label{eq:ProbConstr}
\end{equation}
where $c$ is the indicator variable over inner condition as we will shortly see, $\phi$ is the general belief dependent operator, $\delta$ and $0 \leq \epsilon < 1$ are scalars. In this problem the utility conforms to \eqref{eq:Utility}.  Importantly, as in our previous  paper \cite{Zhitnikov22arxiv}  $\delta$  and $\epsilon$ are supplied by the user. 
 
The constraint $c(b_{k:k+L};\phi, \delta)$ in \eqref{eq:ProbConstr} 
can be of  two forms. The first (cumulative) form is
\begin{align} 
c(b_{k:k+L};\phi,\delta) \bydef \mathbf{1}_{\left\{\left(\sum_{t=k}^{k+L-1} \phi(b_{t+1}, b_{t})\right) >  \delta\right\}}, \label{eq:InnerConstr1} 
\end{align}
and the second (multiplicative) is
\begin{align} 
c(b_{k:k+L};\phi, \delta) \bydef \textstyle\prod_{t=k}^{k+L-1} \mathbf{1}_{\left\{\phi(b_{t+1}, b_{t}) \geq \delta \right\}}. \label{eq:InnerConstr2}
\end{align} 
Further, let us refer to the inner inequality as the inner constraint and correspondingly the outer inequality \eqref{eq:ProbConstr} as the probabilistic (outer) constraint.
From now on, for clarity, let us denote \emph{constraining} return and the \emph{actual} return operator as 
$s(b_{k:k+L};\phi)  \bydef \sum_{t=k}^{k+L-1} \phi(b_{t+1}, b_{t})$ and 
	$s(b_{k:k+L};\rho)  \bydef \sum_{t=k}^{k+L-1} \rho(b_{t+1}, b_{t})$, 
respectively.
To include both cases $\rho$ and $\phi$ in further discussion, we will denote $s(b_{k:k+L};\cdot)$. 

Now, we contemplate what will happen, if $\delta$ is a free parameter a not pre-determined as before. In this case we would like to select action sequence corresponding to largest maximal feasible return (actual or constraining $s(b_{k:k+L};\cdot)$) with probability of at least $1-\epsilon$. That is, maximal $\delta$ yielding that, at most, a single action sequence is feasible.  With this insight in mind, we arrive to our {\bf second} problem of {\bf maximal feasible return} defined as follows 
\begin{align}
	a^* \in \arg\max_{a_{k+} \in \mathcal{A}} \  \underbrace{\mathrm{VAR}_{\epsilon}\big(s(b_{k:k+L};\cdot)|b_k, a_{k+}\big)}_{\obj(b_k, a_{k+})}, \label{eq:Problem2}
\end{align}
where the  Value at Risk (VaR) reads 
\begin{equation}
\begin{split}
	&\mathrm{VaR}_{\epsilon}\bigg(s(b_{k:k+L};\cdot)|b_k,a_{k+}  \bigg) \bydef \delta^* = \\
	&\mathrm{sup}\Big\{\delta: \prob\Big(s(b_{k:k+L};\cdot) > \delta |b_k, a_{k+}\Big) \geq 1- \epsilon \Big\}. 
\end{split}	
\label{eq:VaR}
\end{equation}

Due to noncompliance to Bellman form of \eqref{eq:VaR} computing \eqref{eq:Problem2} is notoriously challenging.

Another way to introduce a belief-dependent constraint to POMDP setting would be by averaging with respect to observations. Namely, the probabilistic constraint in \eqref{eq:ProbConstr} is replaced by  $\mathcal{C}(b_k, a_{k+})\geq \delta$ given by
\begin{align}
 &\!\!\!\mathcal{C}(b_k, a_{k+}) = \underset{\beta_{(k+1)+}}{\mathbb{E}} \Big[ \mathcal{C}^{\beta_{(k+1)+}}(b_k, a_{k+}) \Big| b_k, a_{k+} \Big], \label{eq:OuterExpectedConstr} 
 \end{align}
where
\begin{align}
 &\mathcal{C}^{\beta_{(k+1)+}}(b_k, a_{k+})=  \label{eq:InnerExpectedConstr}\\
 &\!\!\underset{z_{(k+1)+}}{\mathbb{E}} \! \Bigg[\! \sum_{t=k}^{k+L-1}\!\!\!\phi(b_t,a_t, {z}_{t+1},  b_{t+1}) \Big| b_k,\! a_{k+},\! \beta_{(k+1)+} \Big]\! \Big| b_k, \!a_{k+} \Bigg]. \nonumber
\end{align}
If $\phi$ is selected to be Information Gain, \eqref{eq:InnerExpectedConstr} is known as Mutual Information (MI). However, if one transfers the utility \eqref{eq:Utility} to the constraint, in other words, when the  $\rho (\cdot) \equiv \phi(\cdot)$  and we use expectation in \eqref{eq:OuterExpectedConstr} and in \eqref{eq:Utility} such a constraint appears to be problematic. If $\mathcal{U}(\cdot) \equiv \mathcal{C}(\cdot)$,  we can always maximize the utility and ask if optimal utility is larger than $\delta$ ($\obj^{*} > \delta$). In general this is the question of what one verifies first, \emph{optimality} or  \emph{feasibility}.  As we shall further see, in some cases the order does matter and we can save time by  fast feasibility check and cancellation of action sequences.  Another option would be to use a maximum likely sequence of observations $z^{\mathrm{ML}}_{k+1:k+L}$ and check 
\begin{align}
	\Bigg(\sum_{\ell=k}^{k+L-1} \phi(b_{t},a_{t}, z^{\mathrm{ML}}_{t+1},  b_{t+1}) \Bigg) \geq \delta, 
\end{align}
where the maximum likelihood  observation $z^{\mathrm{ML}}_{i+1}$ is obtained as follows. We start from a maximum likely state 
\begin{align}
x^{\mathrm{ML}}_{t+1} \in \argmax_{x_{t+1}} \probd(x_{t+1}|b_t, a_t),	
\end{align} 
and then deterministically draw $\beta_{t+1}$, using \eqref{eq:ConfigModel}. This, in turn, results in 
\begin{align}
z^{\mathrm{ML}}_{t+1} \in \argmax_{z_{t+1}} \probd(z_{t+1}|x^{\mathrm{ML}}_{t+1}, \beta_{t+1})  \label{eq:MaximumLikelihood} 
\end{align}
We can interpret the difference of expected constraint \eqref{eq:InnerExpectedConstr} and our probabilistic risk aware constraint \eqref{eq:ProbConstr} as follows. The conventional constraint is not aware of the distribution of the cumulative values of operator $\phi$. It decides either the constraint is fulfilled or not solely using the expected value. It is possible that the expected value of the constraint fails to represent adequately the underlying distribution. In contrast, our formulation is {\bf distribution aware}. 

In reality to evaluate the probabilistic constraint we shall marginalize over observation sequences to calculate 
\begin{equation}
	\begin{split}
		& \prob (c(b_{k:k+L}; \phi, \delta) | b_k, a_{k+}) = \\
		& \int_{z_{(k+1)+}} \!\!\!\!\!\!\!\!\!\!\!\!\!\! \prob (c(b_{k:k+L}; \phi, \delta)| b_k, a_{k+}, z_{(k+1)+}) \cdot  \\
		&\probd(z_{(k+1)+}|b_{k}, a_{k+})\mathrm{d} z_{(k+1)+}.
	\end{split}
	\label{eq:ProbConstrTheoretical}
\end{equation}

In the following sections, we develop a universal theory to evaluate our proposed probabilistic inequality $\prob (c(b_{k:k+L}; \phi, \delta)=1;\delta | b_k, a_{k+}) \geq 1-\epsilon$
 \emph{adaptively}. On top of that, we expedite the evaluation process even more by extending \emph{simplification} paradigm to our setting, enjoying the substantially improved celerity versus baseline approaches.    

\subsection{Adaptive Belief Tree} \label{sec:AdaptiveBeliefTree}
The integral in eq. \eqref{eq:ProbConstrTheoretical} is not accessible in a general setting. One way to approximately evaluate the \eqref{eq:ProbConstrTheoretical}  is to sample from observation likelihood $\probd(z_{(k+1)+}|b_{k}, a_{k+})$. We assume that we have a fixed budget $m$ of samples of observation laces. Our aim is to leverage the fact that we have a particular structure of the probabilistic condition  \eqref{eq:ProbConstrTheoretical} and to address its evaluation while constructing the belief tree, thereby saving valuable running time or providing a more accurate solution. For clarity, we recite some statements from our previous paper. We remind the reader that this paper focuses on the open-loop setting, namely static action sequences, instead of policies.

Imagine a candidate action sequence $a_{k:k+L-1}$. 
To approximate the utility and the probabilistic constraint (Eq.~\eqref{eq:ProbConstr}), an  online algorithm at the root (for each candidate action sequence) expands upon termination $m$ laces appropriate to the drawn observations 
\begin{align}
\{z^l_{k+1:k+L}\}_{l=1}^{m}. \label{eq:SetObservations}
\end{align} 
Through the paper we label the laces in the belief tree by the superscript $l$ (Fig.~\ref{fig:BeliefTreeBeta}). Each lace $l$ corresponds to a particular realization of the sequence of the beliefs, return $s(b_{k:k+L};\rho)$ or constraining return $s(b_{k:k+L};\phi)$.  The sample approximation of probabilistic constraint \eqref{eq:ProbConstrTheoretical} is
\begin{align}
\frac{1}{m} \sum_{l=1}^{m} c(b^{l}_{k:k+L};\phi, \delta) \geq 1- \epsilon.  \label{eq:ProbConstrSampleApprox}
\end{align}
We employ an already expanded part of the belief tree to bound the expression of the probabilistic constraint from each end using the following adaptive upper and lower bounds
\begin{align}
& \underbrace{\frac{1}{m} \sum_{q=1}^{\tilde{m}} c(b^{l^{q}}_{k:k+L}; \phi, \delta)}_{\mathrm{lb}^{(1)}} \leq  \frac{1}{m} \sum_{l=1}^{m} c(b^{l}_{k:k+L}; \phi, \delta).  \label{eq:Lower}
\end{align}
\begin{align}
& \frac{1}{m}\!\sum_{l=1}^{m} \! c(b^{l}_{k:k+L};\phi, \delta) \!\leq \!\underbrace{\frac{m\!-\!\tilde{m}}{m}\!+\! \frac{1}{m}\!\sum_{q=1}^{\tilde{m}} \! c(b^{l^{q}}_{k:k+L};\phi, \delta)}_{\mathrm{ub}^{(1)}}  \label{eq:Upper}
\end{align}
where, the algorithm already expanded $\tilde{m} \leq m$ laces  in some order. We denote expanded laces by a sub-sequence $q \in 1 \ldots \tilde{m}$, such that $l^q$ is the index of the observation sequence, i.e, $l^q \in 1\dots m$. By the adaptivity we mean the expanding lowest number of laces depending on the situation to accept or discard the candidate action sequence.  
\begin{figure}[t]
	\centering
	\includegraphics[width=0.7\columnwidth]{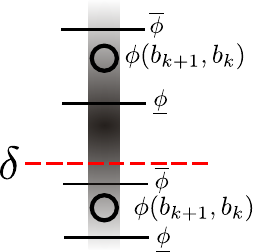}
	\caption{Conceptual visualization of our \emph{simplification} approach (Section \ref{sec:AdaptiveSimpl}). For clarity we show a myopic setting. The gradient displays the probability density, i.e., a larger number of samples lands in the area of greater intensity. Using the bounds, we want to assess whether the fraction of the observation laces above $\delta$ is at least $1-\epsilon$. As we see, we can invalidate bottom sample $\phi$ using solely the upper bound $\overline{\phi}$. In a similar manner, we can validate the upper sample $\phi$ using solely the lower bound $\underline{\phi}$. Note that the width of the vertical strip has no role in this visualization. }
	\label{fig:Simpl}
\end{figure}   

\subsection{Adaptive Simplified Constraint Evaluation} \label{sec:AdaptiveSimpl}
As introduced in  
\cite{Zhitnikov22ai}, \cite{Sztyglic21arxiv_b}, \cite{Elimelech22ijrr}, \cite{Shienman22icra}, the simplification paradigm  seeks to ease the computational burden in the decision making problem while providing performance guarantees. The latter is achieved by applying bounds over various quantities in the decision making problem (e.g.~bounds over a reward function). In this section we extend this concept to our   probabilistic belief-dependent constrained POMDP setting of Eq.~\eqref{eq:ProbConstr}.

Suppose we have adaptive deterministic bounds over $\phi$, i.e.~these  bounds hold for any realization of the beliefs. Further, evaluating these bounds is computationally cheaper than the operator $\phi$. 
Let us present the main theorem of this section, which will shed light on how these bounds can be utilized, propagating their adaptivity further to the adaptive constraint evaluation.    
\begin{thm}[Simplification machinery] \label{thm:SimplSample}
	Imagine a sample set  of the observations laces $\{z^q_{k+1:k+L}\}_{q=1}^{n}$. Assume that  
	$\forall q$  holds
	\begin{align}
	\underline{\phi}(b^q_{\ell+1}, b^q_{\ell}) \leq \phi(b^q_{\ell+1}, b^q_{\ell}) \leq \overline{\phi}(b^q_{\ell+1}, b^q_{\ell}). \label{eq:BoundsSample}
	\end{align}	 
	Let two forms of sample version of inner constraint \eqref{eq:InnerConstr1} and \eqref{eq:InnerConstr2} variants with bounds be
	\begin{align} 
	&\overline{c}(b^q_{k:k+L};\overline{\phi}, \delta) \bydef \mathbf{1}_{\left\{\left(\sum_{t=k}^{k+L-1} \overline{\phi}(b^q_{t+1}, b^q_{t})\right) >  \delta\right\}}, \\
	&\underline{c}(b^q_{k:k+L}; \underline{\phi}, \delta) \bydef \mathbf{1}_{\left\{\left(\sum_{t=k}^{k+L-1} \underline{\phi}(b^q_{t+1}, b^q_{t})\right) >  \delta\right\}},  
	\end{align}
	and 
	\begin{align} 
	&\overline{c}(b^q_{k:k+L}; \overline{\phi},\delta) \bydef \textstyle\prod_{t=k}^{k+L-1} \mathbf{1}_{\left\{\overline{\phi}(b^q_{t+1}, b^q_{t}) \geq \delta \right\}}, \\
	&\underline{c}(b^q_{k:k+L}; \underline{\phi}, \delta) \bydef \textstyle\prod_{t=k}^{k+L-1} \mathbf{1}_{\left\{\underline{\phi}(b^q_{t+1}, b^q_{t}) \geq \delta \right\}}.  
	\end{align}
	Eq.~\eqref{eq:BoundsSample}, in turn, implies that the following inequalities are satisfied without dependency on the form
	\begin{align}
	&\!\!\!\!\sum_{q=1}^{\tilde{m}} \underline{c}(b^q_{k:k+L}; \underline{\phi},\delta)\!\! \leq \!\!  \sum_{q=1}^{\tilde{m}} c(b^q_{k:k+L};\phi,\delta) \!   \leq \!  \sum_{q=1}^{\tilde{m}} \overline{c}(b^q_{k:k+L}; \overline{\phi},\delta).\!\! \label{eq:LowerUpperSimplSample} 
	\end{align} 	
\end{thm}
We provide a detailed proof of Theorem~\ref{thm:SimplSample} in  Appendix \ref{proof:SimplSample}. 

Let us now show how to speed up the process of evaluation of
the probabilistic constraint from \eqref{eq:ProbConstr}. The key component of the
acceleration is that the adaptivity of the bounds is delegated to
adaptivity of the probabilistic constraint bounds \eqref{eq:LowerUpperSimplSample}. Assume
the bounds from \eqref{eq:BoundsSample} are adaptive, using insights provided by Theorem~\ref{thm:SimplSample}, we first check if 
\begin{align}
\frac{1}{m} \sum_{q=1}^{\tilde{m}}\underline{c}(b^{l^q}_{k:k+L}; \underline{\phi},\delta) \overbrace{\geq}^{?} 1-\epsilon. \label{eq:LowerCheckSimpl}
\end{align} 
If the above relation holds we declare that the outer constraint is fulfilled. If not, we probe if 
\begin{align}
\frac{1}{m} \sum_{q=1}^{\tilde{m}}\overline{c}(b^{l^q}_{k:k+L}; \overline{\phi},\delta)  \overbrace{<}^{?} 1-\epsilon. \label{eq:UpperCheckSimpl}
\end{align} 
If yes, we declare that the outer constraint is violated. In case we are not able to say anything (both relations do not hold), we tighten the bounds. In other words, we make the bounds closer to the actual value of $\phi$ (e.g., by utilizing more particles \cite{Zhitnikov22ai}, \cite{Sztyglic22iros} or mixture belief components \cite{Shienman22icra}). 
We presented a conceptual visualization of our simplification approach in Fig.~\ref{fig:Simpl}. 



Now our goal is to merge the insights gained in section \ref{sec:AdaptiveBeliefTree} with the simplification. Clearly from \eqref{eq:Lower} and the left side  of \eqref{eq:LowerUpperSimplSample}  we have that 
\begin{align}
&1-\epsilon \overbrace{\leq}^{?} \underbrace{\frac{1}{m} \sum_{q=1}^{\tilde{m}} \underline{c}(b^{l^{q}}_{k:k+L}; \underline{\phi},\delta)}_{\mathrm{lb}^{(2)}} \leq  \underbrace{\frac{1}{m} \sum_{q=1}^{\tilde{m}} c(b^{l^{q}}_{k:k+L};\phi, \delta)}_{\mathrm{lb}^{(1)}}. \label{eq:MergedLowerCheck}
\end{align}
Similarly from \eqref{eq:Upper} and right side of \eqref{eq:LowerUpperSimplSample} holds
\begin{equation}
 \label{eq:MergedUpperCheck}
\begin{split}
&\underbrace{\!\frac{m\!-\!\tilde{m}}{m}  +\frac{1}{m} \sum_{q=1}^{\tilde{m}} c(b^{l^{q}}_{k:k+L};\phi, \delta)}_{\mathrm{ub}^{(1)}}\! \leq \\
&\! \underbrace{\leq \!\frac{m\!-\!\tilde{m}}{m}  +\frac{1}{m} \sum_{q=1}^{\tilde{m}} \overline{c}(b^{l^{q}}_{k:k+L};\overline{\phi},\delta)}_{\mathrm{ub}^{(2)}}\overbrace{<}^{?} 1\!-\!\epsilon\!
\end{split}
\end{equation}
\begin{figure}[t] 
	\centering
	\begin{minipage}[t]{0.7\columnwidth}
		\centering
		\includegraphics[width=\columnwidth]{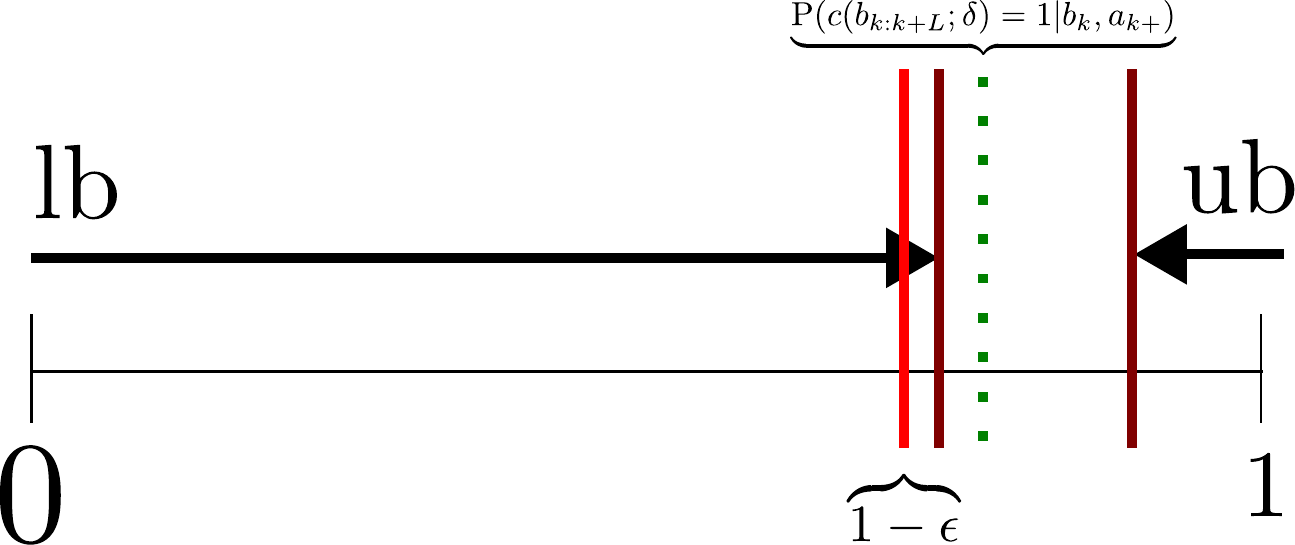}
		\subcaption{}
		\label{fig:ChallengingAdapt}
	\end{minipage}%
	\vfill
	\begin{minipage}[t]{0.7\columnwidth}
		\centering
		\includegraphics[width=\columnwidth]{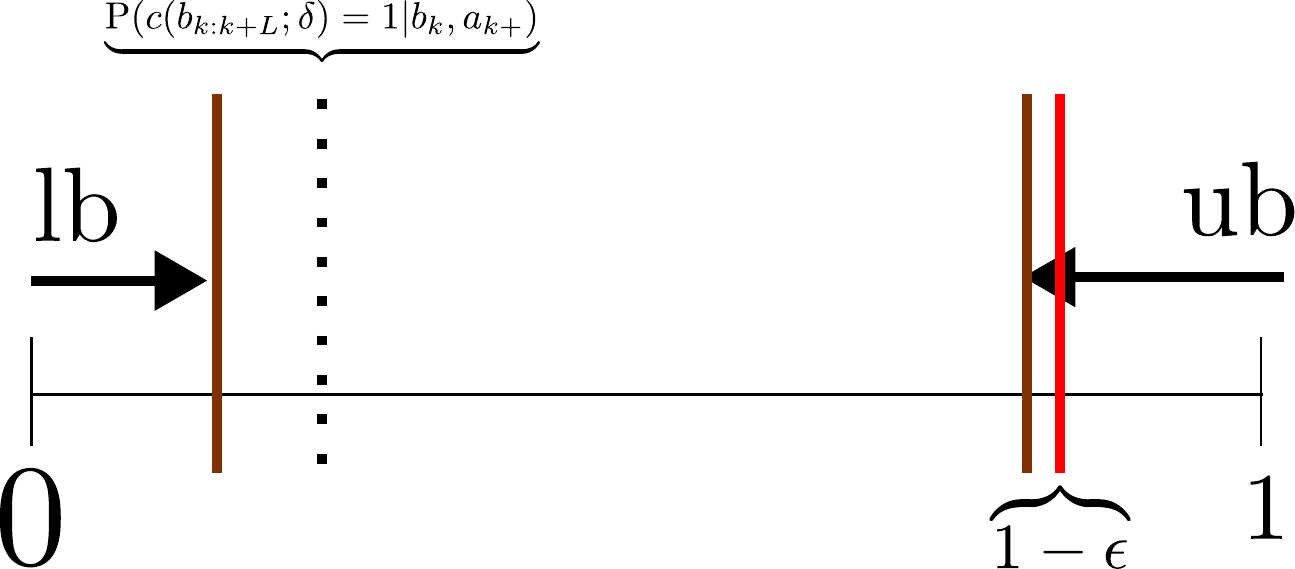}
		\subcaption{}
		\label{fig:EasyAdapt}
	\end{minipage} 
	\vfill
	\begin{minipage}[t]{0.7\columnwidth}
		\centering
		\includegraphics[width=\columnwidth]{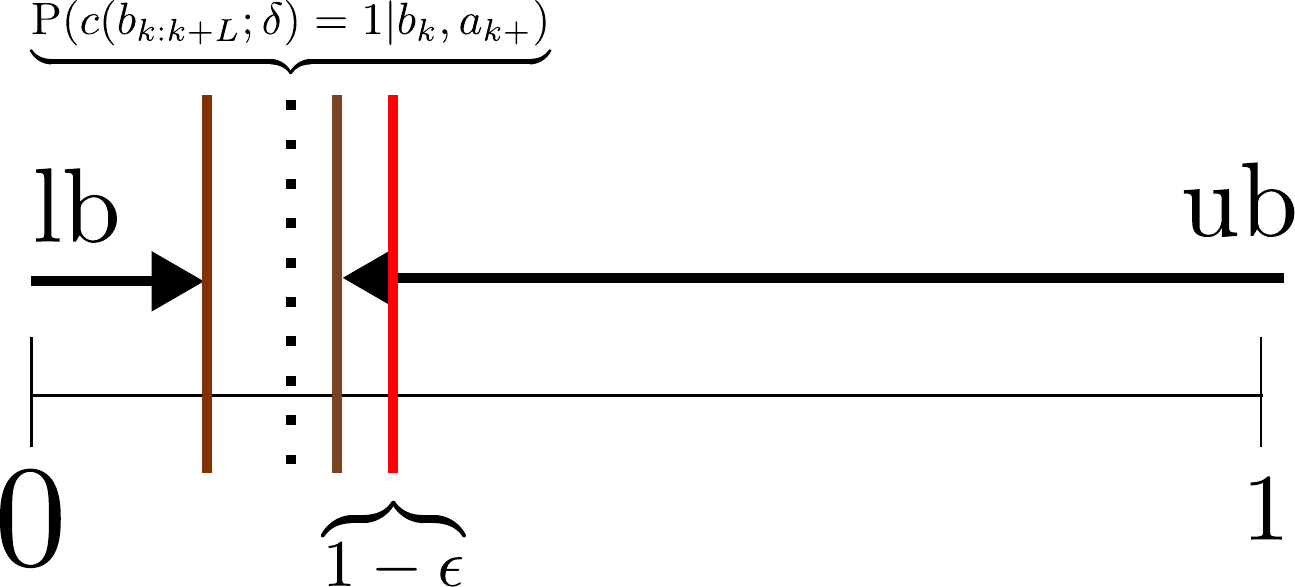}
		\subcaption{}
		\label{fig:ChallengingAdapt2}
	\end{minipage}
	\vfill
	\begin{minipage}[t]{0.7\columnwidth}
		\centering
		\includegraphics[width=\columnwidth]{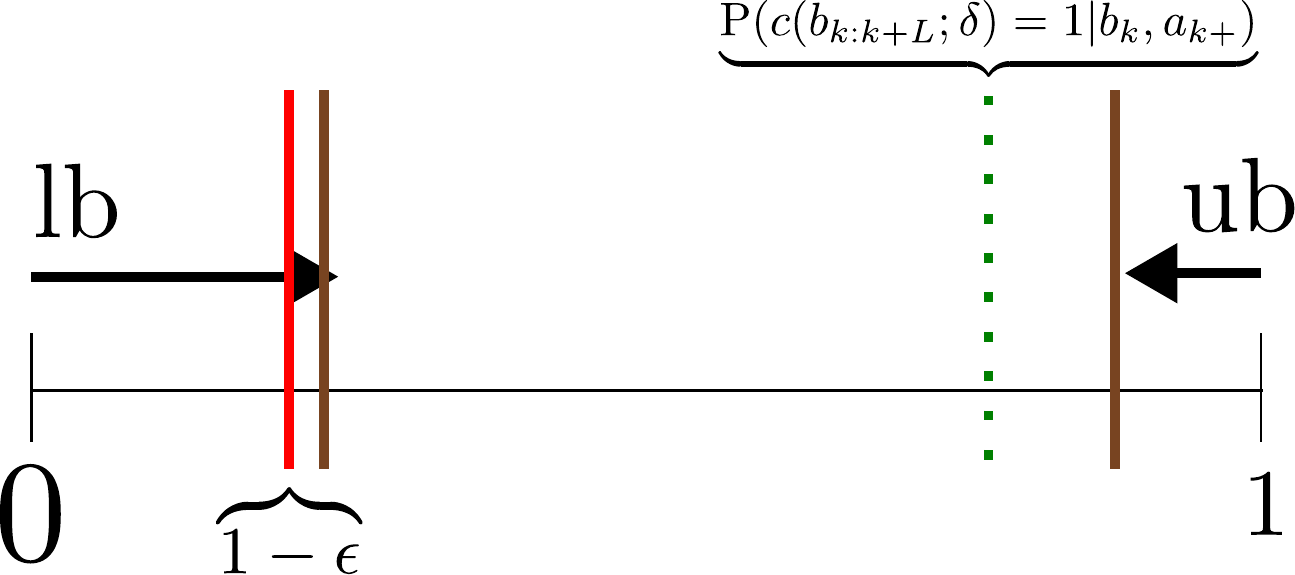}
		\subcaption{}
		\label{fig:EasyAdapt2}
	\end{minipage}
	\caption{Visualization of adaptation from section \ref{sec:AdaptSimplTwoLayers}. Note in all scenarios the value of dashed line is {\bf unknown}. The red line represents the confidence level $1-\epsilon$ to be satisfied with probabilistic constraint. \textbf{(a)} Conceptual illustration of challenging scenario. To {\bf accept} such an action the lower bound shall go a long way; \textbf{(b)} Conceptual illustration of easy scenario, with a few contractions of the upper bound, the action is {\bf discarded}; \textbf{(c)} Another interesting situation, here the upper bound shall go a long way to {\bf discard} an action. \textbf{(d)} With a few shrinkage iterations the lower bounds accepts the action sequence. } 
	\label{fig:BoundsProperties}
\end{figure}
By a question mark, we denote the inequalities that shall be fulfilled online to check whether the outer constraint is met \eqref{eq:MergedLowerCheck} or violated \eqref{eq:MergedUpperCheck}. If we cannot incur the status of the outer constraint we shall add more laces (adapt the first layer bound) or/and tighten the bounds from \eqref{eq:BoundsSample}. Such an approach permits adaptive evaluation of the outer constraint in Eq.~\eqref{eq:ProbConstr} before expanding the $m$ laces of the belief sequences $b_{k:k+L}$.

\subsection{The adaptation} \label{sec:AdaptSimplTwoLayers}
It occurs that the proposed bounds have riveting properties. To describe a pair of lower ($\mathrm{lb}^{(1)},\mathrm{lb}^{(2)}$) and a pair of upper bounds ($\mathrm{ub}^{(1)},\mathrm{ub}^{(2)}$) simultaneously, we omit the superscript.
The lower bound is bounded by zero $0\leq \mathrm{lb}$ and the upper bound is bounded by one $ \mathrm{ub} \geq 1$. When we adapt the bounds, we add at most a single lace to the appropriate sum. Therefore, the step of adaptation of the bounds is $\nicefrac{1}{m}$.
 
When we expand a single lace $\tilde{m} \leftarrow \tilde{m}+1$, the lower bound makes a step if $c(b^l_{k:k+L};\phi, \delta)=1$, otherwise the upper bound makes a step if $c(b^l_{k:k+L};\phi, \delta)=0$. Alternatively, when we increase the simplification level, some already expanded laces possibly switch from $0$ to $1$ ($\underline{c}(b^{l^{q}}_{k:k+L};\underline{\phi},\delta)$ for some $l^q$), contracting the lower bound, and some from $1$ to $0$ ($\overline{c}(b^{l^{q}}_{k:k+L};\overline{\phi},\delta)$ for some $l^q$) , tightening the upper bound. 

Importantly when we expand a single observation lace and calculate $\underline{c}(b^{l^{q}}_{k:k+L};\underline{\phi},\delta)$  we will obtain one with probability {\bf at most}  $\prob (c(b_{k:k+L}; \phi, \delta)=1 | b_k, a_{k+})$.  Similarly we will obtain $\overline{c}(b^{l^{q}}_{k:k+L};\overline{\phi},\delta) = 0$ at the new expanded lace with probability at most $\prob (c(b_{k:k+L}; \phi, \delta)=0 | b_k, a_{k+})$.
Both these probabilities are not accessible.

Further, we have four scenarios illustrated in Fig.~\ref{fig:BoundsProperties}. By analyzing these scenarios, we can speculate about anticipated speedup.  In Fig.~\ref{fig:BoundsProperties} we show by the red vertical line several positions of the outer threshold $1-\epsilon$  from \eqref{eq:ProbConstr}. The first scenario, shown in Fig.~\ref{fig:ChallengingAdapt}, is challenging. The unavailable to us probabilistic constraint is fulfilled (shown by green dashed vertical line  in Fig.~\ref{fig:ChallengingAdapt}); therefore, no matter how many iterations we perform, invalidation using the calculated $\mathrm{ub}$ and Eq.~\eqref{eq:MergedUpperCheck} is not possible; only validation using $\mathrm{lb}$ and \eqref{eq:MergedLowerCheck} will eventually be possible. As we observe, many contractions of the $\mathrm{lb}$ would be required, as we see in Fig.~\ref{fig:ChallengingAdapt} up until $\mathrm{lb}$ becomes larger  than $1-\epsilon$ according to Eq.~\eqref{eq:MergedLowerCheck}. Conversely, if with a {\bf large margin} the outer constraint is violated as we see in Fig.~\ref{fig:EasyAdapt}, we discard the action sequence  with a few tightening iterations using $\mathrm{ub}$ and \eqref{eq:MergedUpperCheck}. We contemplate a similar behavior in reciprocal cases  (Figs.~\ref{fig:ChallengingAdapt2} and \ref{fig:EasyAdapt2}). To conclude the adaptation can be challenging in cases described in Figs.~\ref{fig:ChallengingAdapt} and \ref{fig:ChallengingAdapt2}.

The fact that we have a pair of lower ($\mathrm{lb}^{(1)},\mathrm{lb}^{(2)}$) and a pair of upper bounds ($\mathrm{ub}^{(1)},\mathrm{ub}^{(2)}$)  raises the question which bound from each pair shall we adapt in case that a pair is inconclusive.

When we cannot incur whether the outer constraint from \eqref{eq:ProbConstrSampleApprox} is fulfilled, we shall decide to refine the bounds or add more laces (observation sequences). Luckily for us, these two operations are parallelizable in terms of multithreading. We simultaneously refine the simplification levels, as in \cite{Sztyglic21arxiv_b} of the bounds, and add more laces up until the decision is possible.    

To conclude this section, we proposed a two-layered approach to ease a computational burden. The first layer expresses adaptivity in terms of the number of observation laces. The second layer permits utilization of the adaptive deterministic bounds on realizations of $\phi|b_k, \pi$.  

One example of using our technique is to save time in open loop planning or spend more time on the action sequences which fulfill the probabilistic constraint.  
With such an approach, we are able to cut down on the cost of exhaustively validating candidate action sequences. In the setting of Gaussian high dimensional beliefs in SLAM problem, a lower number of observation laces is especially important due to loop closures. 
Another example is the closed loop setting, where we deal with policies. 


Thus far we presented general theory, and now we specifically address the second problem described by Eq.~\eqref{eq:Problem2}.
\subsection{Maximal Feasible Return} \label{sec:Bisection}
\begin{figure}
	\centering
	\includegraphics[width=0.7\columnwidth]{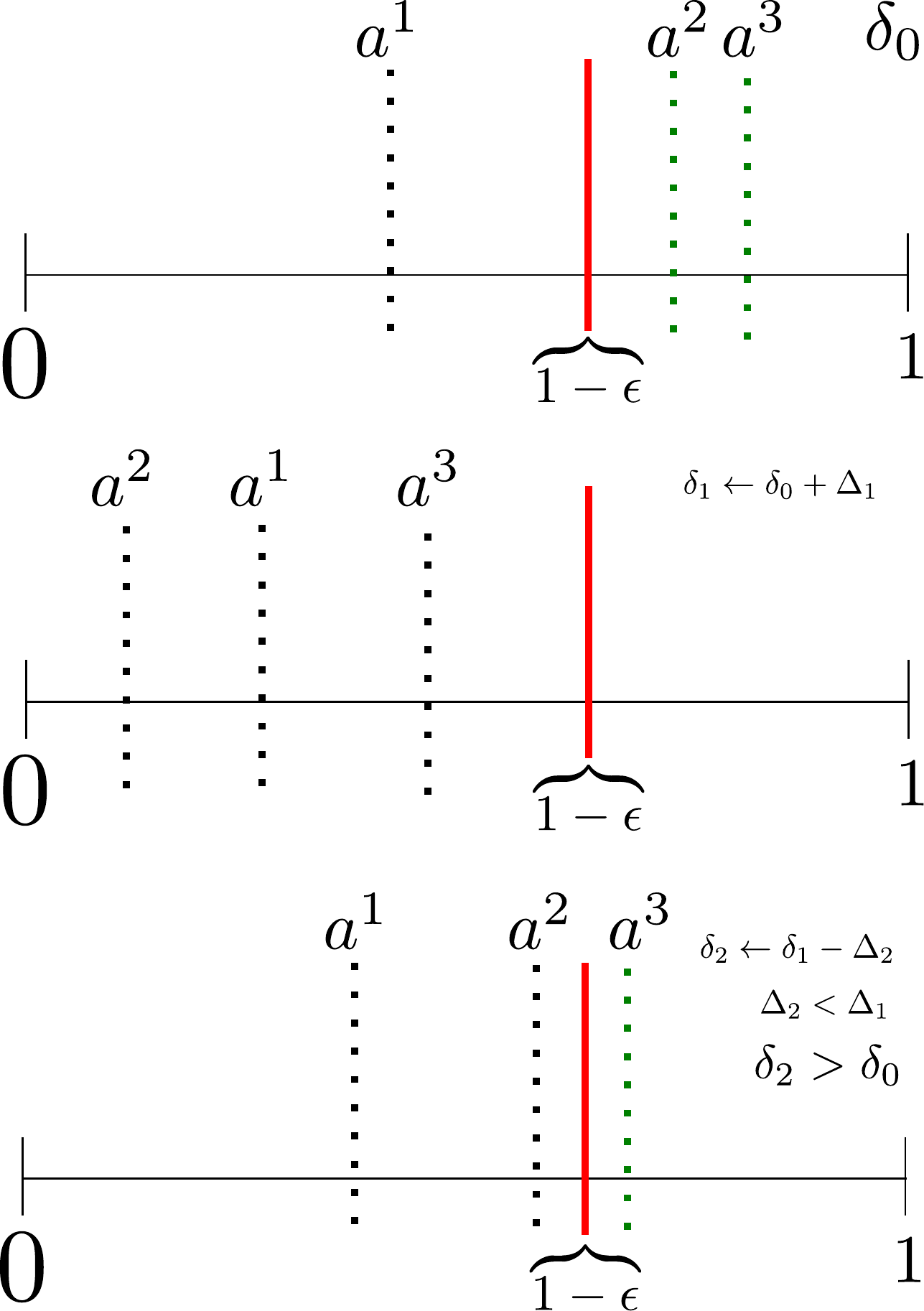}
	\caption{Visualization of Alg.~\ref{alg:PurelyConstrained}. We never increase the step size. Therefore, as we see, each candidate action sequence in the bottom visualization is shifted to the left relative to the situation displayed in the top. The action sequence $a^1$ can be safely discarded in the top illustration (Section \ref{sec:Bisection}). }
	\label{fig:Bisection}
\end{figure}
Picture in your mind that you guess the $\delta$ and the step size  $\Delta$. For clarity we drop the dependence of $s$ on $b_{k:k+L}$. However, we shall remember that a single realization of $s$ corresponds to a single lace in the belief tree (Fig.~\ref{fig:BeliefTreeBeta}).  Observe the following pair of relations
\begin{align}
	&\prob\big(s >  \delta |b_k, a_{k+}\big) \geq  \prob\big(s  >  \delta + \Delta  |b_k, a_{k+}\big),  \\
	&\prob\big(s  >  \delta |b_k, a_{k+}\big)  \leq   \prob\big(s  >  \delta  -  \Delta |b_k, a_{k+}\big).
\end{align}
Suppose we fulfill the probabilistic inequality with $\delta$ for a subset of candidate action sequences, that is, $\prob\big(s > \delta |b_k,a_{k+}\big) \geq 1-\epsilon$. We shall increase $\delta$ to invalidate more candidate action sequences up until a single  candidate action sequence is left. Currently invalidated candidate action sequences can be discarded for eternity, they will never fulfill the outer constraint with larger $\delta$, due to the never increasing step size in our approach of alternating increases and decreases of $\delta$. Now, suppose we violate the probabilistic inequality with $\delta$, that is, $ 1-\epsilon > \prob\big(s > \delta |b_k, a_{k+}\big) $ for all the candidate action sequences. We shall decrease the $\delta$ to render more candidate action sequences feasible. If we will obtain $\delta$ such that all the candidate action sequences besides the single one are invalidated, we know that this candidate action sequence maximizes \eqref{eq:VaR}.  
This is the underlying principle of Alg.~\ref{alg:PurelyConstrained}. See visualization in Fig.~\ref{fig:Bisection}.  As we see in Fig.~\ref{fig:Bisection},  $\delta_2 > \delta_0$ so $\prob\big(s \! > \! \delta_0 |b_k, a_{k+}\big)\! \geq \! \prob\big(s \! > \! \delta_2  |b_k, a_{k+}\big)$. To the step size, we imply the bisection principle. 
To rephrase it, we solve    
\begin{equation}
\begin{split}
&a_{k+}^*, \delta^{*} = \argmax_{\{a_{k+}\}} \ \max_{\delta} \quad  \delta  \\ 
&\text{s.t.   } \exists \ a_{k+} \in \mathcal{A} \ : \ \prob(c(b_{k:k+L};\phi, \delta)|b_k, a_{k+}) \geq 1-\epsilon\\
\end{split}
\end{equation}
This formulation is equivalent to solving the maximal feasible return problem portrayed by equation \eqref{eq:Problem2}.
Before we continue, note that in Appendix~\ref{app:SampleApprox} we discuss sample approximations used in our proposed algorithms.   
We are ready for the next section, where we formulate algorithms to tackle both of our problems.

\subsection{Algorithms} \label{sec:Algorithms}

\input{./floats/Alg.tex}

\input{./floats/AlgBruteForce.tex}

\input{./floats/Alg_purely_constr.tex}

\input{./floats/Alg_baseline.tex}

In this section, we present four algorithms. All the algorithms receive as input the set of candidate action sequences. How these action sequences are obtained is out of the scope of this paper. For both our problems, we propose our technique and describe the baseline. 
Our algorithms shall surpass the baseline methods in terms of celerity or/and quality of the solution. 
Importantly, the overhead from the adaptation shall be neglectable.

\subsubsection{Optimality under probabilistic constraint}
For the first problem \eqref{eq:ProbConstr}, we adaptively check the feasibility of all the action sequences and select the optimal from the set of feasible action sequences Alg.~\ref{alg:StochasticExploration}. The competing approach is finding the optimal action sequence and verifying feasibility afterwards, see  Alg.~\ref{alg:Baseline1}. 

\subsubsection{Maximal feasible gain}
Here, we propose our adaptive method described in Section \ref{sec:Bisection} and summarized in  Alg.~\ref{alg:PurelyConstrained} and evaluate/compare it  versus the brute force maximization of Value at Risk Alg.~\ref{alg:Baseline2}.  

Having introduced the algorithms we shall discuss possible drawbacks and overhead.  
\subsection{Adaptation overhead}
When we use presented above adaptation mechanisms, we store laces $c^{l}(b_{k:k+L};\phi, \delta)$ for every expanded $l$. Accordingly, the memory consumption is elevated, however not much since these are boolean values. Moreover, we shall evaluate the inner constraint and perform the sum for multiple values of $\delta$ in Alg.~\ref{alg:PurelyConstrained}. Nevertheless, as we believed and verified by the experiments, this overhead is neglectable compared to the saved time on skipped laces due to loop closures. In addition, these additional operations can be easily parallelized in terms of multithreading. 

We can, however, encounter a worst-case scenario. Imagine the $\epsilon$ is close to $1$   from the left. Many action sequences will satisfy the probabilistic constraint. In general, we can say that a more accurate precision of $\delta$ will be required to differentiate between the action sequences since the working area is closer to zero and the interval $[0, 1-\epsilon)$ is shorter. Therefore, more iterations of  Alg.~\ref{alg:PurelyConstrained} will be required. Moreover, a pair of action sequences may be extremely close to each other in terms of Value at Risk, requiring a tremendous amount of iterations of the Alg.~\ref{alg:PurelyConstrained}. To solve this issue, we shall introduce a final precision. 

\section{Application to Belief Space Planning} \label{sec:Application}
In this section we apply proposed algorithms to informative planning with high dimensional robot's state. We express the exploration problem with our framework \eqref{eq:ProbConstr}.  
\subsection{Belief structure}\label{sec:BeliefStructure}
Let us delve into the mechanics of maintaining and updating high-dimensional belief on top of a stochastic process - sequential decision making. 
A standard and widely used tool used to maintain a high-dimensional belief is a factor graph \cite{Koller09book}. Its building blocks are the probabilistic motion and observation models. These models induce probabilistic dependencies over the state variables. The models are the factors that comprise the factor graph.

In this paper, the stochastic motion and observation models are described by the following dependencies involving the Gaussian distributed sources of stochasticity. 
\begin{align}
	&x_{i+1} = f(x_t, a_t; w_t), \quad w_t \sim \mathcal{N}(0, W_t),  \label{eq:StructuredMotion} \\
	&z^{j^{\nu}}_{t} = g(x_t, \ell^{j^{\nu}}; v_t) , \quad v_k \sim \mathcal{N}(0, V_t), \label{eq:StructuredObservation}
\end{align}
where $W_t$ and $V_t$ are covariance matrices.
In this paper we assume that the data association is solved. Namely, in general, the belief would be (see, e.g.,\cite{Pathak18ijrr}, \cite{Tchuiev19iros})
\begin{align}
	&\!\!\!\!\!b_k(\boldsymbol{x}_k) \bydef \probd(\boldsymbol{x}_k|b_0, a_{0:k-1}, z_{1:k} ) =  \\
	&\!\!\!\!\!\sum_{\beta_{1:k}} \probd(\boldsymbol{x}_k|b_0, a_{0:k-1},z_{1:k}, \beta_{1:k} )\prob(\beta_{1:k}|b_0, a_{0:k-1}, z_{1:k}). \nonumber
\end{align} 
We, however, assume that given an observation the the realization of corresponding $\beta$ is known. This fact simplifies the belief structure as such
\begin{align}
	&\probd(\boldsymbol{x}_k|b_0, a_{0:k-1}, z_{1:k} ) = \probd(\boldsymbol{x}_k|b_0, a_{0:k-1}, z_{1:k}, {\color{red}{\beta_{1:k}}} ).
\end{align} 
Applying the Bayes Rule, we arrive at 
\begin{align}
	&b_k(\boldsymbol{x}_k)\! \propto \!  b_0(x_0) \prod_{i=1}^k \! \Bigg( \! \underbrace{\probd_T\big(x_t \big|  x_{t-1}, a_{t-1}\big)}_{\text{motion factor}}\! \prod_{\nu_{i}=1}^{n(\beta_{i})} \!  \underbrace{\probd_Z\big(z^{\nu_i}_t \big| x_t,\ell^{j^{\nu_i}}\big)}_{\substack{\text{observation} \\ \text{factor}}} \Bigg) \label{eq:FactorGraph}
\end{align} 
Eq.~\eqref{eq:FactorGraph} can be illustrated as a factor graph \cite{Dellaert17foundations}. In this paper, we utilize Gaussian probabilistic models \eqref{eq:StructuredMotion} and \eqref{eq:StructuredObservation} in our simulations. 
\subsection{Information Gain} 
Let us address the cumulative form of the inner constraint \eqref{eq:InnerConstr1}.   
Similar to \cite{Placed22arxiv}, we define the operator $\phi$ as follows 
\begin{align}
	\phi(b',b) \bydef \mathrm{IG}(b,a,z', b'). \label{eq:ConstraintOperator}
\end{align}
There are various ways to define the Information Gain (IG) over a pair of the beliefs. 
One possibility is as follows   
\begin{align}
	\mathrm{IG}(b,a,z', b') = - \mathrm{h}(b') +  \mathrm{h}(b).
\end{align}
The differential entropy $\mathrm{h}(b)$ is defined by 
\begin{align}
	\mathrm{h}(b) \bydef \int_{x} b(x)\log b(x) \mathrm{d} x. \label{eq:Entropy}
\end{align}

To employ Alg.~\ref{alg:PurelyConstrained} we require to supply minimal ($\delta^{\mathrm{min}}$) and maximal ($\delta^{\mathrm{max}}$) threshold for inner constraint (eq. \eqref{eq:InnerConstr1} and \eqref{eq:InnerConstr2}). Let us unveil how we do that.

Differential entropy \eqref{eq:Entropy} was widely researched in the context of multivariate Gaussian beliefs and led to the formulation of the $D$-optimality criterion being the multiplication of eigenvalues of the covariance matrix of the belief (the volume of $d$-dimensional parallelepiped proportional to the volume of a hyper-ellipse manifested by the covariance matrix). The information gain becomes 
\begin{align}
	0 \underbrace{\leq}_{ \substack{ \mathrm{require}  \\  I' \geq I}} \underbrace{-\sqrt[d]{\prod_i^d \lambda^{\prime,i}}}_{I^{\prime}\leq 0} + \underbrace{\sqrt[d]{\prod_i^d \lambda^{i}}}_{-I} \leq \sqrt[d]{\prod_i^d \lambda^{i}}, \label{eq:DoptGain}
\end{align}  
where $d$ is the dimension of the subset of the variables selected from the Gaussian belief.  
For the reason that our focus is on the uncertainty of the environment surrounding the robot, we select {\bf all the landmarks} as such a subset alongside the {\bf current robot pose}. Since we do not add landmarks in the planning session, the same dimensionality is preserved.

Moreover from \eqref{eq:DoptGain} we elicit that the maximal feasible $\delta$ is $\delta^{\mathrm{max}} \bydef \sqrt[d]{\prod_i^d \lambda^{d}}$.  Meaning, the uncertainty has been reduced to zero in the resulting  Gaussian (partial) belief.   To continue exploration we select $\delta^{\mathrm{min}} = 0$. To summarize in the setting of multivariate Gaussian beliefs
$0 < \delta \leq  \sqrt[d]{\prod_i^d \lambda^{i}}.$

Having untangled these aspects, we are keen to demonstrate the superiority of the proposed approach in the following section.

\section{Simulations and Results} \label{sec:Sim}
\begin{figure}[t] 
	\centering
	\begin{minipage}[t]{0.49\columnwidth}
		\includegraphics[width=\columnwidth]{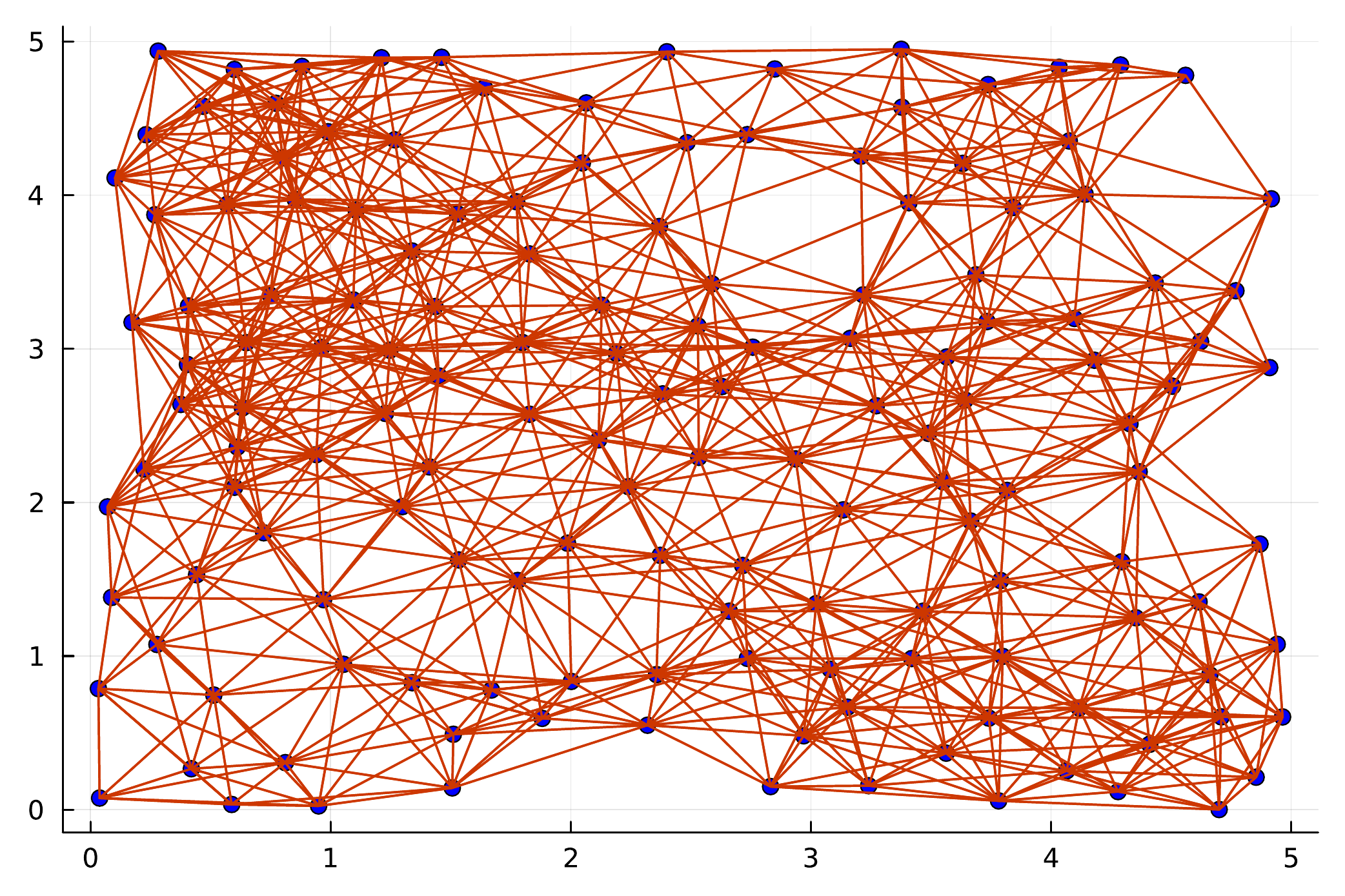}
		\subcaption{PRM}
		\label{fig:PRMPaths}
	\end{minipage}%
	\hfill
	\begin{minipage}[t]{0.49\columnwidth}
		\centering 
		\includegraphics[width=\textwidth]{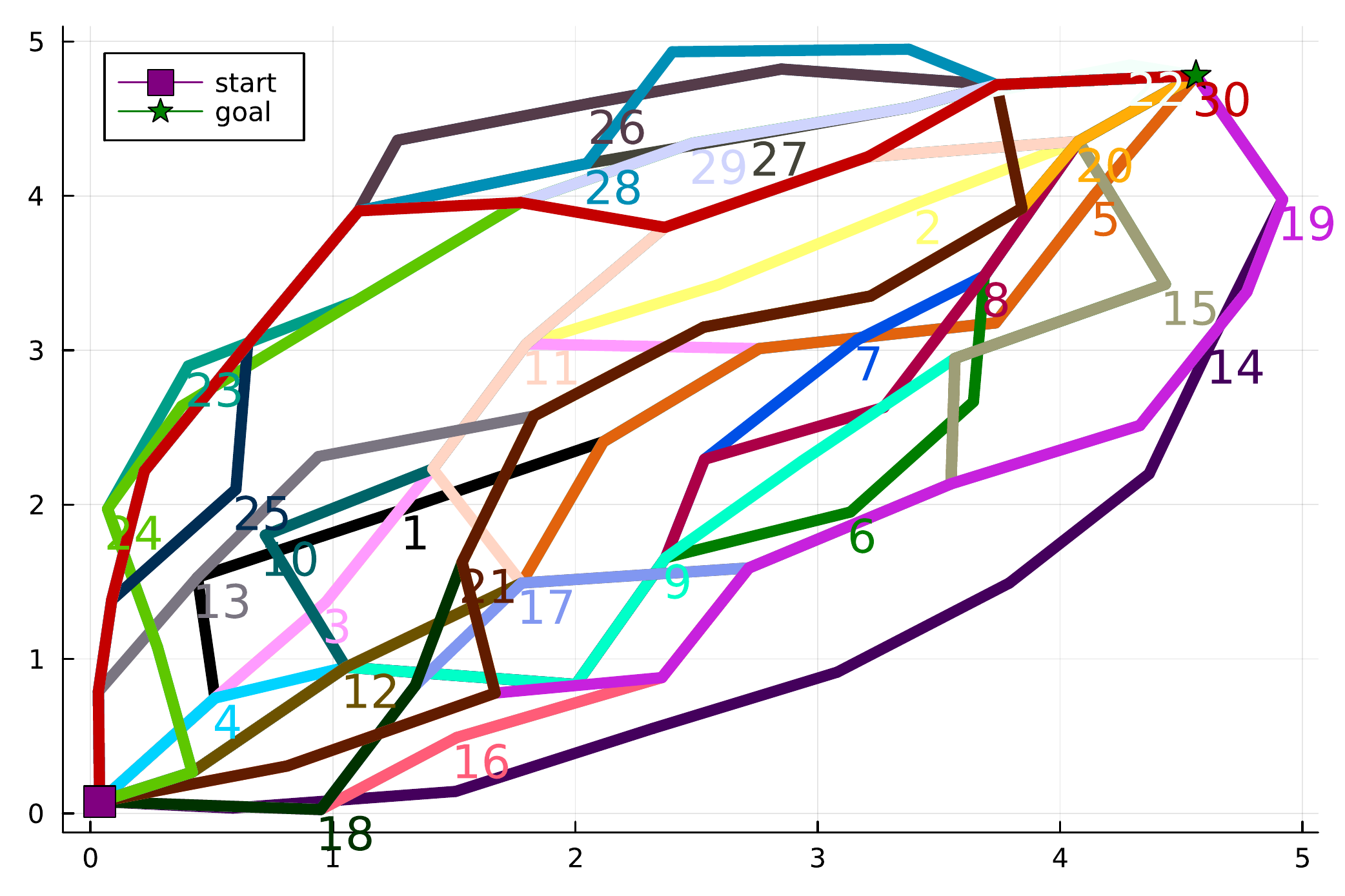}
		\subcaption{Obtained diverse paths}
		\label{fig:paths}
	\end{minipage} 
	\caption{\label{fig:PRMandPaths}Separate, algorithmically selected paths to the goal \textbf{(b)} on top of PRM \textbf{(a)}. We show the path number on the vertex, which is removed for finding the subsequent diverse path. The last's path number is shown at its final vertex (the goal).}
\end{figure}

\begin{figure}[t]

\end{figure}   
The previous discussion leads us to the actual implementation and simulations of the proposed methods in section \ref{sec:Algorithms}. We evaluate our approach by tackling the problem of navigation to the goal in {\bf unknown environments} as an incarnation of BSP. 
The simulation involves a {\bf {highly realistic}} SLAM scenario using the gtsam library \cite{Dellaert12tr}. 

As  previously mentioned, the generation of candidate paths is not the focus of this paper. Therefore, we generate candidate paths following a similar procedure to \cite{Indelman17arj}. First, we employ a well-studied Probabilistic Road MAP (PRM) method \cite{Kavraki96tra}. Then, on top of PRM, to obtain diverse shortest paths, we remove a single vertex from the previous path and utilize Breadth-First Search on the reduced PRM. The path generation requires only the boundaries of an unknown map.  In such a way, we obtain $|\mathcal{A}|$ diverse paths to the goal of various lengths. These paths constitute the space of action sequences $\mathcal{A}$ (Fig.~\ref{fig:PRMPaths}). To avoid confusion, we recite that any other method for generating candidate paths would be applicable to evaluate our proposed techniques.
We illustrated the described above in Fig.~\ref{fig:PRMandPaths}. 

To keep the examination clear, we do not perform re-planning sessions. Instead, we have a preliminary mapping session with manually supplied to the robot action sequence of unit length motion primitives. In the preliminary session, the robot starts from $b_0$, detects the landmarks, incorporates them into its state, and obtains the belief $b_k$. This belief serves as input to the planning session on top of the candidate paths. After a single planning session, the robot follows the calculated optimal path.

We assume Gaussian sources of stochasticity. For motion \eqref{eq:StructuredMotion} and observation \eqref{eq:StructuredObservation} models we select  $W_t = \|a\|_2 \cdot \mathrm{diag}(0.015 , 0.015, 0.015)$ and $V_t = \mathrm{diag}(0.001 , 0.001)$ respectively. Noticeable, we need to multiply the motion model covariance matrix by the action length since our actions are of variable length.  Our prior belief is Gaussain over the robot's pose  $b_0 \sim \mathcal{N}(\mu_0, \Sigma_0)$ with the parameters $\mu_0 = (0.0, 0.0, 0.0)^T$, $\Sigma_0 = \mathrm{diag}(0.001 , 0.001, 0.001)$. The boundaries of our map are $[0, 5] \times [0,5]$.

We utilize the popular incremental solver ISAM2 \cite{Kaess12ijrr} to maintain the belief. Noticeably,  loop closures impose a computational challenge even with such a sophisticated incremental solver. Especially since we need to perform inference for each posterior node in the constructed belief tree. This fact makes early eliminating or accepting actions highly important for efficient robot's operation. 

The robot constructs a belief tree for each candidate path withing planning session. With each promotion of the depth of the belief tree, we reduce the number of observations at each belief node by factor two, up to a possible single observation at the lowest levels. Once the maximal number of observations of the belief node is expanded, we maintain a circular slider that selects the subsequent observation with the following arrival at this belief node. 

The advantage of our proposed methods is acceleration without compromising the solution quality. We calculate the speedup using the following equation 
\begin{align}
	\frac{t^{\mathrm{baseline}}-t^{\mathrm{our}}}{t^{\mathrm{baseline}}}. \label{eq:speedup}
\end{align}
Each planning session is initialized by the same seed.  In addition, we do the same calculation to the relative fraction of the  skipped laces
\begin{align}
\frac{n^{\mathrm{total}}-n^{\mathrm{expanded}}}{n^{\mathrm{total}}}. \label{eq:LacesFrac}
\end{align}
\begin{figure}[t]
	\centering
	\includegraphics[width=\columnwidth]{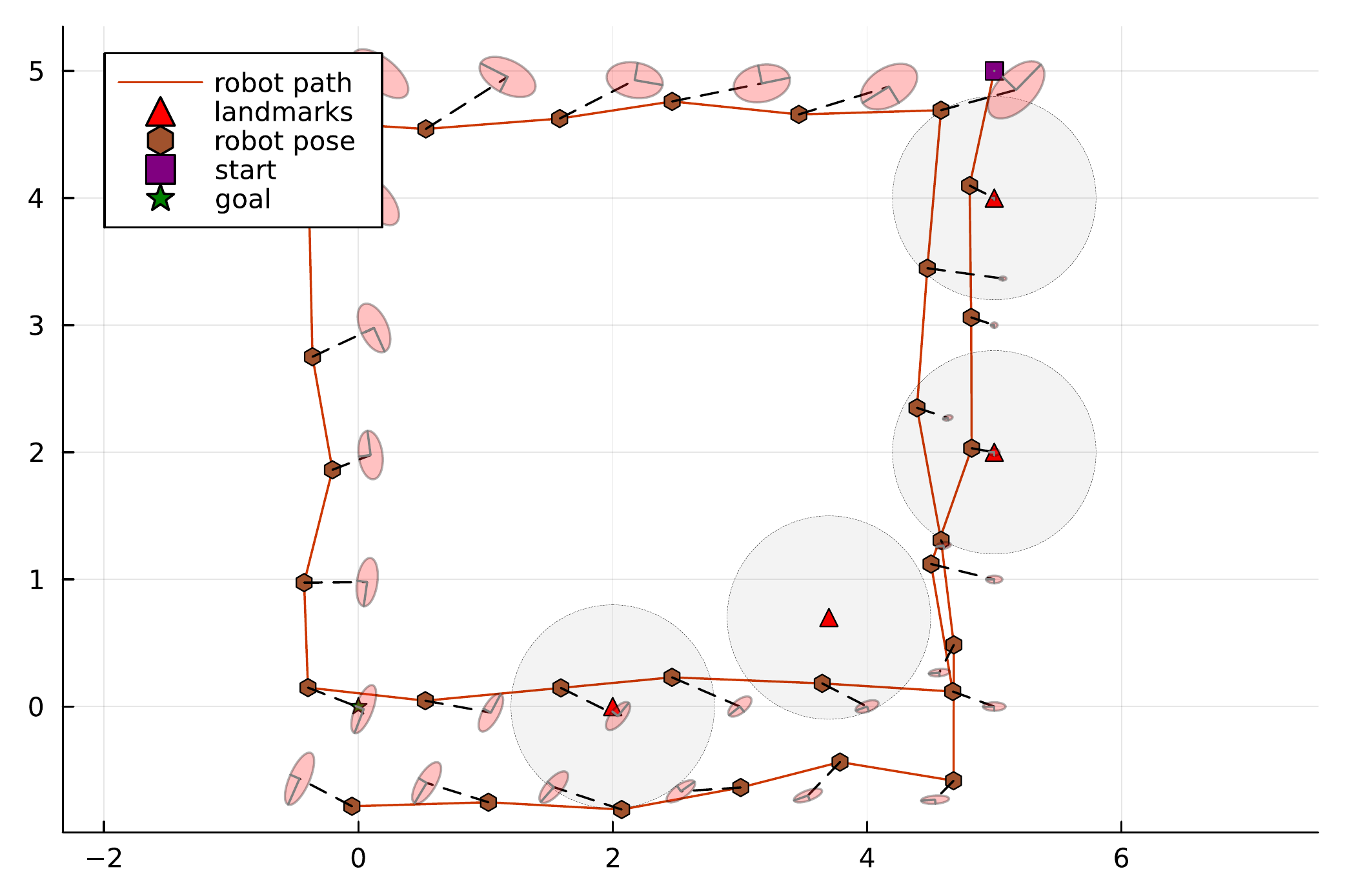}
	\caption{Robot first preliminary mapping session, by transparent gray circles, we depict landmarks' visibility radius. The robot starts at the top right corner and moves towards the bottom left corner making two full squares. As we can see, the robot passed inside the visibility radius of the landmarks, detected them and incorporated them to its state. We show covariance ellipses for current robot poses. The landmarks visibility radius is $0.8$. By the dashed line we connect estimated robot pose with ground truth. }
	\label{fig:warmup1}
\end{figure}   
Note that maximal values of \eqref{eq:speedup} and \eqref{eq:LacesFrac} are $1$. This means that our approach skipped all the laces ($n^{\mathrm{expanded}}=0$ in \eqref{eq:LacesFrac}) and run in zero time ($t^{\mathrm{our}}=0$ in \eqref{eq:speedup}). 
\subsection{Optimality under Probabilistic  Constraint}
\input{./floats/tableOptimalityUnderProbConstraint.tex}
\begin{figure}[t] 
	\centering
	\includegraphics[width=\columnwidth]{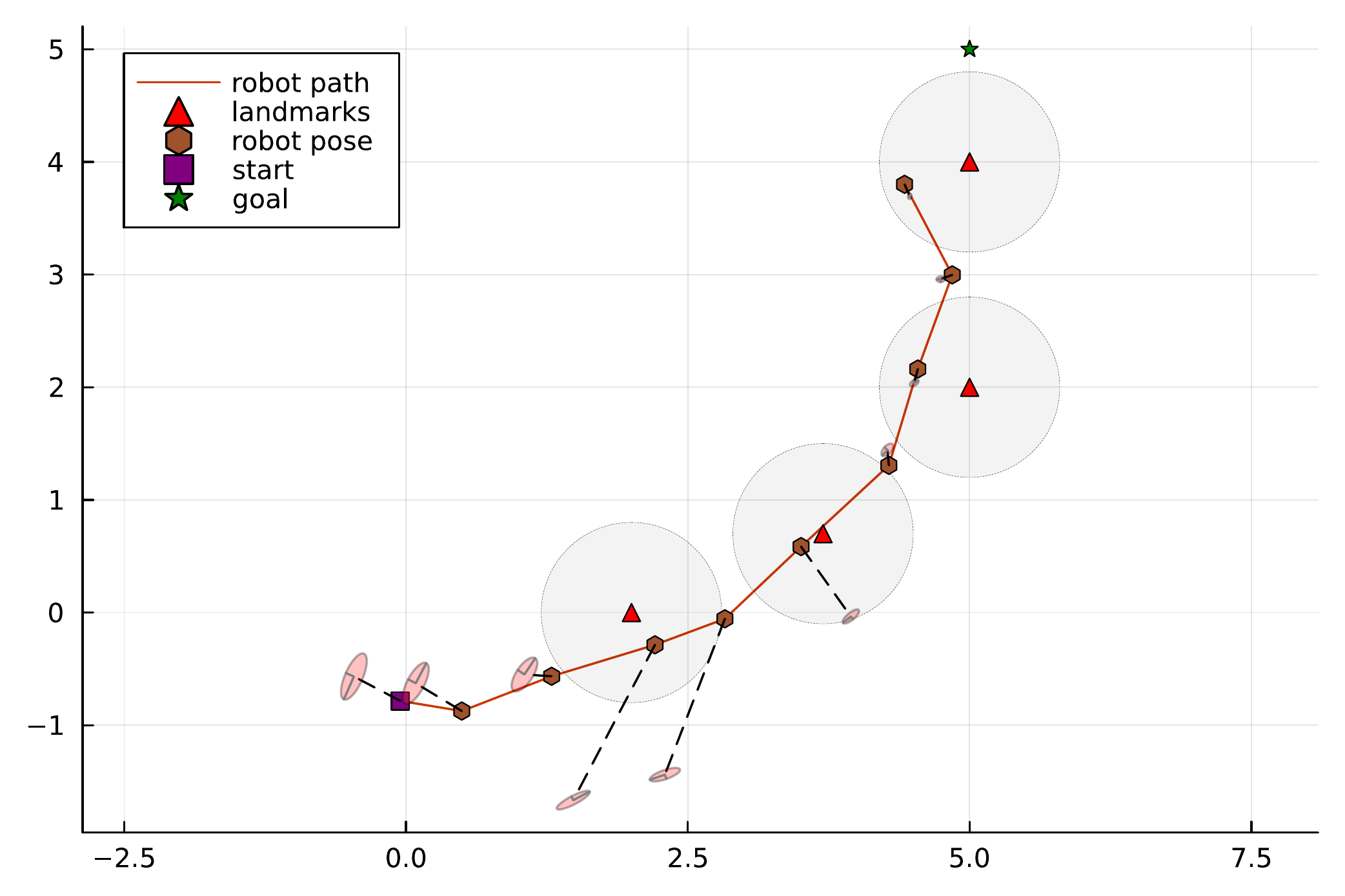}
	\caption{ Alg.~\ref{alg:Baseline1} and Alg.~\ref{alg:StochasticExploration} both selected path number $14$ from Fig.~\ref{fig:paths} as optimal. We recognize that a pair of landmarks nearest to starting position ($5, 5$) of preliminary mapping session in Fig.~\ref{fig:warmup1} greatly contribute to uncertainty diminishment since the robot twice made a loopclosure there.}
	\label{fig:robotMoving}
\end{figure}
Following the previous discussion, we continue with the experiments. 
We start from our first problem \eqref{eq:ProbConstr} (optimality under a probabilistic constraint) and study Alg.~\ref{alg:StochasticExploration} versus Alg.~\ref{alg:Baseline1}. Noticeable, in Alg.~\ref{alg:Baseline1} we do not have a mechanism for early action dismissing until we expand all the observation laces per action sequence.  We examine a simple scenario with four landmarks. We show the preliminary mapping session in Fig.~\ref{fig:warmup1}. We elicit that, as anticipated, the uncertainty over the belief grows until the robot makes a full square and starts to experience loop closures. The path number $14$ is highly likely to be optimal from an information perspective since the belief is Gaussian, and this path lies closest to the landmarks. We employ Alg.~\ref{alg:StochasticExploration} with $m=300$ laces per path from Fig.~\ref{fig:PRMPaths}, $\delta=0.0$ and various values of $\epsilon$. Our resolution in terms of $\epsilon$ is $\Delta^{\epsilon}=\nicefrac{1}{m}$. Empirically we found that for $\epsilon \in [0, 0.023]$,  without dependency on $m$ as expected, all the paths were discarded as unfeasible ($7$ from $300$ laces given path $14$ were violating the inner constraint). Meaning no path is present with the fraction of the laces larger than $1-0.023$ fulfilling inner constraint. 

We show a rigorous comparison versus Alg.~\ref{alg:Baseline1} in Table.~\ref{tbl:OptimalityUnderProbConstraint}. In Fig.~\ref{fig:robotMoving} we display the robot following the identified optimal path. Note that with Alg.~\ref{alg:StochasticExploration} we do not accelerate decision making when we cannot discard  action sequences. We shall note that due to internal gtsam multi-threading,  measuring the time speedup is a challenging task. To alleviate that we repeat each run in Table.~\ref{tbl:OptimalityUnderProbConstraint} five times and report averaged running time and the speedup obtained from averaged value of the running time. Remarkably, from the bottom line of Table~\ref{tbl:OptimalityUnderProbConstraint} we observe that with extremely loose probabilistic constraint ($\epsilon = 0.9$) we do not eliminate any action sequence but the running time is not larger than the baseline. This fact indicate that there is no overhead from adaptation.
\begin{figure}[t]
	\centering
	\includegraphics[width=\columnwidth]{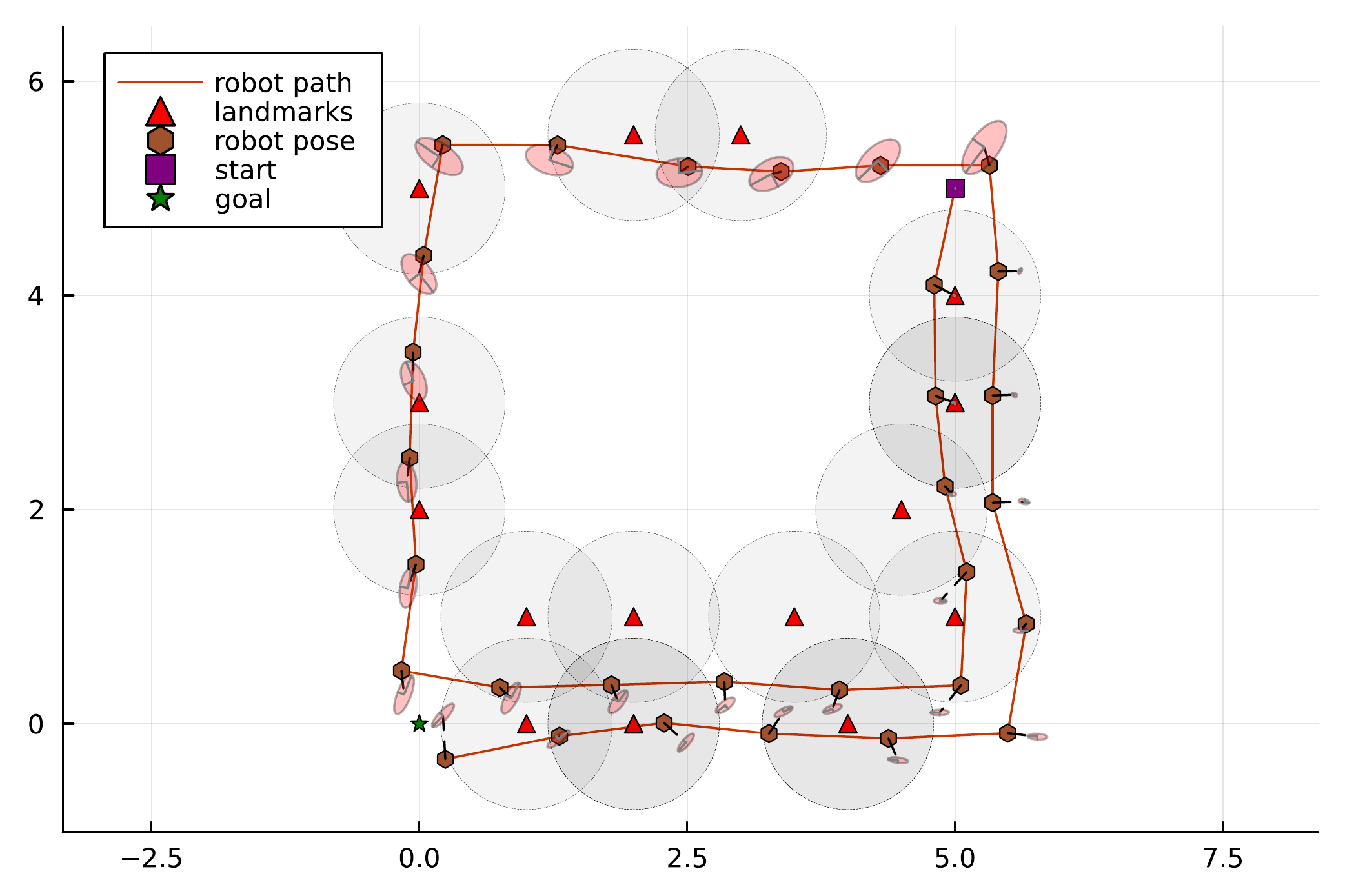}
	\caption{Robot second preliminary mapping session, by transparent gray circles we depict landmarks' visibility radius. As we can see that robot detected the landmarks and incorporated to its state. The landmarks visibility radius is $0.8$.  }
	\label{fig:warmup2}
\end{figure}  
\begin{figure}[t] 
	\centering
	\includegraphics[width=\columnwidth]{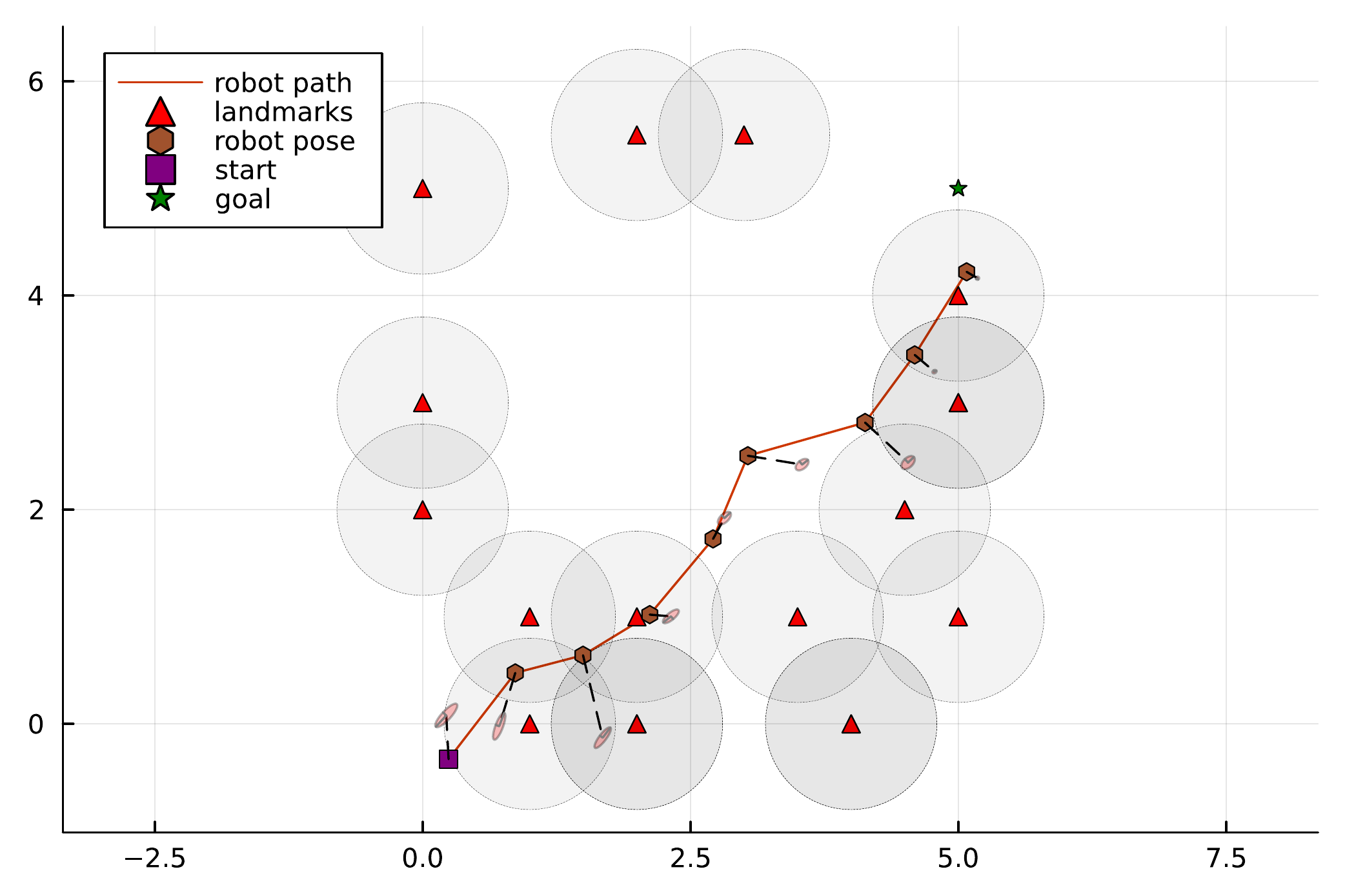}
	\caption{ Alg.~\ref{alg:Baseline2} and Alg.~\ref{alg:PurelyConstrained} both selected path number $4$ from Fig.~\ref{fig:paths} as optimal.}
	\label{fig:robotMoving2}
\end{figure}
\subsection{Maximal Feasible Return}
\begin{figure}[t] 
	\centering
	\includegraphics[width=\columnwidth]{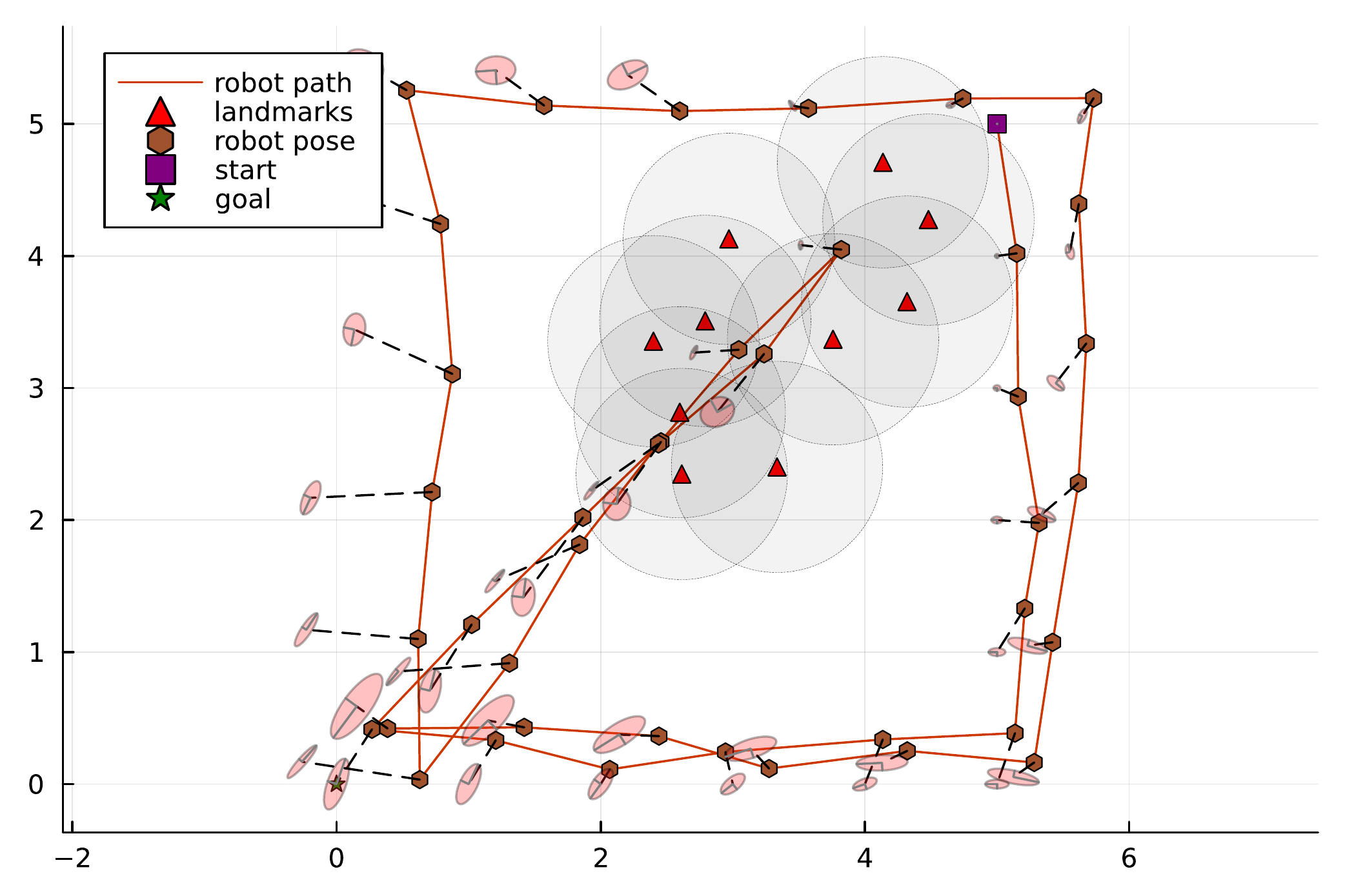}
	\caption{ Illustration of the third preliminary mapping session with randomly drawn landmarks.}
	\label{fig:robotWarmUp3}
\end{figure}
We continue to our second problem  (maximal feasible return \eqref{eq:Problem2}).  Let us increase the number of landmarks to obtain more good candidate paths for information gathering. We show our second preliminary mapping session in Fig.~\ref{fig:warmup2}. Here we need many paths with nonegative information gain to examine using Alg.~\ref{alg:PurelyConstrained} early acceptance as well and not only early invalidation as was done in previous section. Our baseline is Alg.~\ref{alg:Baseline2},  which calculates VaR in a straightforward way. We report results in Table.~\ref{tbl:MaximalFeasibleReturn}. 

We also have an additional simulation with randomly drawing landmarks.   For gtsam stability purposes we add random landmarks uniformly on the square $[2, 5]\times[2,5]$. We also slightly changed the preliminary action sequence (Fig.~\ref{fig:robotWarmUp3}). Results are presented in Table.~\ref{tbl:statistical}.  In our simulations we set the final precision to $1\cdot 10^{-6} \cdot \delta^{\mathrm{max}}$. As we witness from Tables~\ref{tbl:MaximalFeasibleReturn} and \ref{tbl:statistical} we {\bf always} obtain a significant speedup. However, early action elimination appears to be more prominent than early accept. We explain it as follows. It is more likely that will be paths violating the probabilistic constraint and we land at the scenario depicted in Fig.~\ref{fig:EasyAdapt}. Conversely, it is less likely that many paths fulfill the probabilistic constraint and we will land at the scenario depicted in Fig.~\ref{fig:EasyAdapt2}.
\input{./floats/table.tex}
\input{./floats/table_statistical.tex}
\subsection{Technical details}
We used 3 computers with the following characteristics: 
\begin{enumerate}
	\item  8 cores Intel(R) Xeon(R) CPU E5-1620 v4 working at 3.50GHz with 80 GB of RAM;
	\item  8 cores Intel(R) Xeon(R) CPU E5-1620 v4 working at 3.50GHz with 64 GB of RAM;
	\item  16 cores  11th Gen Intel(R) Core(TM) i9-11900K working 3.50GHz with 64 GB of RAM.
\end{enumerate}
\section{Conclusions} \label{sec:Conclusions}
We presented a novel adaptive technique to evaluate probabilistic belief-dependent constraints. On top of that, we provably extended the simplification paradigm to our setting. Our rigorous theory is summarized by two novel adaptive algorithms solving optimality under a probabilistic constraint problem and maximal feasible return problem correspondingly. Our algorithms return identical quality or more profitable solution in a fraction of the baseline running time. By a more profitable solution, we mean elimination of all candidate action sequences using  Alg.~\ref{alg:StochasticExploration}, thereby preventing the robot from  redundant actions when the robot is already deployed and operates online, e.g, stopping exploration. Extensive simulations show the superiority of our methods. In the exceptionally challenging problem of active SLAM with a high dimensional state, we obtained a typical speedup of $30\%$.   
\begin{appendices}
\section{Theoretical Observation Likelihood} \label{app:ObsLikelihood}
To express the observation in terms of probabilistic models available to our disposal  we marginalize over the $x_{t+1}$
\begin{align}
	&\probd(z_{t+1}|b_t, a_t, \beta_{t+1}) \prob(\beta_{t+1}|b_t, a_t) = \nonumber\\
	&\int_{\boldsymbol{x}_{t+1}}\probd(z_{t+1}|b_t, a_t, \beta_{t+1}, \boldsymbol{x}_{t+1}) \cdot \\
	&\probd(\boldsymbol{x}_{t+1}|b_t, a_t, \beta_{t+1}) \prob(\beta_{t+1}|b_t, a_t)\mathrm{d} \boldsymbol{x}_{t+1} = \nonumber\\
	&\int_{\boldsymbol{x}_{t+1}}\probd(z_{t+1}|b_t, a_t, \beta_{t+1}, \boldsymbol{x}_{t+1}) \cdot \label{eq:ObsLikelihood}\\
	& \probd(\boldsymbol{x}_{t+1}|b_t, a_t) \prob_{\beta}(\beta_{t+1}|x_{t+1})\mathrm{d} \boldsymbol{x}_{t+1}.  \nonumber
\end{align}
All quantities in the \eqref{eq:ObsLikelihood} are available for us. Such a representation enables us to draw the observations in look-ahead step $t+1$. 
\section{Sample approximations} \label{app:SampleApprox}
The core of our sample approximations is sequential sampling the observations $z_{t+1}|b_t, a_{t}, \beta_{t+1}$ using previously sampled $\beta_{t+1}|b_t, a_t$.
Following the theoretical derivation presented in Appendix~\ref{app:ObsLikelihood}, we leverage the structure verified by \eqref{eq:ObsLikelihood} in the following way. First, we sample the last pose and the landmarks from the corresponding marginal of the belief. Since our belief is Gaussian, this operation does not introduce a problem. We just pull the appropriate portion of the covariance matrix and the mean value. 
\begin{align}
	x^o_{t+1} \sim \underbrace{\probd(x_{t+1}, \{\ell^j\}_{j=1}^{M(k)} |b_t, a_t)}_{\text{Gaussian}}.
\end{align}	
Afterwards, we deterministically decide the configuration of visible landmarks using \eqref{eq:ConfigModel} and draw samples of the observation laces \eqref{eq:SetObservations} from the observation model \eqref{eq:TotalObsModel}. Finally, the sample approximation of  $\mathcal{U}$ and  $\mathcal{C}(\cdot)$ are denoted by $\hat{\mathcal{U}}^{(m)}$ and  $\hat{\mathcal{C}}^{(m)}$ respectively and calculated by sample means of $\{s(b^l_{k:k+L})\}_{l=1}^m$. Similarly $\widehat{\mathrm{VaR}}^{(m)}_{\epsilon}$ is obtained by sample quantile. 
\section{Proofs}

\subsection{Proof of Theorem \ref{thm:SimplSample} (Simplification machinery)} \label{proof:SimplSample}
It is sufficient to show, that for every sample $z^q_{k+1:k+L}$ holds 
	\begin{align}
	 \underline{c}(b^q_{k:k+L};\underline{\phi}, \delta) \leq   c(b^q_{k:k+L};\phi, \delta)   \leq  \overline{c}(b^q_{k:k+L}; \overline{\phi}, \delta). \label{eq:toprove}
\end{align} 

We start from the left inequality of \eqref{eq:toprove}. Assume that $\underline{c}(b^q_{k:k+L};\underline{\phi}, \delta) = 0$. This implies that trivially $\underline{c}(b^q_{k:k+L};\underline{\phi}, \delta) \leq   c(b^q_{k:k+L};\phi, 
\delta)$. Now suppose that $\underline{c}(b^q_{k:k+L};\underline{\phi},\delta) = 1$. If the inner constraint of the form \eqref{eq:InnerConstr1} so 
\begin{align}
	\left(\sum_{t=k}^{k+L-1} \phi(b^q_{t+1}, b^q_{t})\right) \geq \left(\sum_{t=k}^{k+L-1} \underline{\phi}(b^q_{t+1}, b^q_{t})\right) >  \delta.
\end{align}
We have that $c(b^q_{k:k+L};\phi,\delta) = 1$.
If the inner constraint of the second form \eqref{eq:InnerConstr2} so $\forall t$
\begin{align}
	 \phi(b^q_{t+1}, b^q_{t}) \geq \underline{\phi}(b^q_{t+1}, b^q_{t}) >  \delta.
\end{align}
Again we arrive at $c(b^q_{k:k+L};\phi,\delta) = 1$. To prove the inverse direction, observe that if $c(b^q_{k:k+L};\phi,\delta) = 0$, we behold the following situation with the first form 
\begin{align}
	\delta \geq \left(\sum_{t=k}^{k+L-1} \phi(b^q_{t+1}, b^q_{t})\right) \geq \left(\sum_{t=k}^{k+L-1} \underline{\phi}(b^q_{t+1}, b^q_{t})\right).
\end{align}
and with the second form  $\exists t$ such that  
\begin{align}
	\delta \geq \phi(b^q_{t+1}, b^q_{t}) \geq \underline{\phi}(b^q_{t+1}, b^q_{t}).
\end{align}
It follows that for both forms we have that $\underline{c}(b^q_{k:k+L};\underline{\phi}, \delta) \leq   c(b^q_{k:k+L};\phi, 
\delta)$.

Now we prove the right inequality of \eqref{eq:toprove}. 
If $\overline{c}(b^q_{k:k+L};\overline{\phi}, \delta) = 1$ it is trivial.  Assume that $\overline{c}(b^q_{k:k+L};\overline{\phi},\delta) = 0$. For the first form  \eqref{eq:InnerConstr1} it means that
\begin{align}
	\left(\sum_{t=k}^{k+L-1} \phi(b^q_{t+1}, b^q_{t})\right) \leq \left(\sum_{t=k}^{k+L-1} \overline{\phi}(b^q_{t+1}, b^q_{t})\right) \leq   \delta.
\end{align}
Subsequently, for the second form \eqref{eq:InnerConstr2}  it means that  $\exists t$ such that 
\begin{align}
	\phi(b^q_{t+1}, b^q_{t}) \leq \overline{\phi}(b^q_{t+1}, b^q_{t}) \leq  \delta.
\end{align}

Arguing in the similar manner as with the lower bound for inverse direction, suppose that $c(b^q_{k:k+L};\phi,\delta) = 1$. We obtain
\begin{align}
	\delta < \left(\sum_{t=k}^{k+L-1} \phi(b^q_{t+1}, b^q_{t})\right) \leq \left(\sum_{t=k}^{k+L-1} \overline{\phi}(b^q_{t+1}, b^q_{t})\right),
\end{align}  
and $\forall t$
\begin{align}
	\delta < \phi(b^q_{t+1}, b^q_{t}) \leq \overline{\phi}(b^q_{t+1}, b^q_{t}).
\end{align}

For both forms we have that   $c(b^q_{k:k+L};\phi, \delta)   \leq  \overline{c}(b^q_{k:k+L}; \overline{\phi}, \delta)$.

This concludes the proof. Note that we also land at a identical result for theoretical counterparts of following probabilities and not sample approximations by taking the limit. 
\begin{align}
&\!\!\!\!\!\lim_{m \to \infty} \frac{1}{m} \sum_{q=1}^{m} \!	\underline{c}(b^q_{k:k+L};\underline{\phi}, \delta)\! \leq \! \lim_{m \to \infty} \frac{1}{m} \sum_{q=1}^m  c(b^q_{k:k+L};\phi,\delta)\!\!\\
&\!\!\!\!\! \lim_{m \to \infty} \frac{1}{m} \sum_{q=1}^{m} \! c(b^q_{k:k+L}; \phi, \delta)\! \leq\! \lim_{m \to \infty} \frac{1}{m} \sum_{q=1}^m  \!\overline{c}(b^q_{k:k+L}; \overline{\phi}, \delta).\!\!\!
\end{align} 
\qed
\end{appendices}

\bibliographystyle{IEEEtran}
\input{paper.bbl}




\end{document}

%% file: packages.tex
\usepackage[utf8]{inputenc}
\usepackage{amsmath}
\usepackage{amssymb}
\usepackage{graphicx}
\usepackage{cancel}
\usepackage{dsfont}
\usepackage{algorithm}
\usepackage{algpseudocode}
\usepackage{tikz}
\usepackage{bbm}
\usepackage{dsfont}
\usepackage{graphicx}
\usepackage{dsfont}
\usepackage{graphicx}
\usepackage{subcaption}
\usepackage{multirow}
\usepackage{placeins}

\usepackage{enumerate}
\usepackage{xspace}
\usepackage{xcolor,colortbl}
\usepackage{tabularx}
\usepackage[labelfont=bf]{caption}

\usepackage{multicol}
\usepackage{multirow}
\usepackage{wrapfig}

%% file: floats/Alg.tex
\begin{algorithm}[t]
	\begin{algorithmic}[1]
		\caption{Optimality under probabilistic constraint}
		\label{alg:StochasticExploration}
		\State {\bf Input:} $\mathcal{A}$ \Comment{Set of the action sequences}
		\State $a^*_{k+} \leftarrow \mathrm{undef}$, $\hat{\obj}^{*}_{(m)} \leftarrow -\infty$, $S \leftarrow \{\}$ 
		\For {each $a_{k+} \in \mathcal{A}$}
		\For { $l = 1 : m $ }
		\State Draw observation sequence $z^l_{k+1:k+L}$ 
		\State Calculate $c(b^l_{k:k+L};\phi, \delta)$, 
		\If {$1-\epsilon \leq \frac{1}{m}\sum_{q=1}^l c(b^q_{k:k+L};\phi, \delta)$} \Comment{Outer constraint is fulfilled}
			\State   $S \leftarrow S \cup a_{k+}$  \Comment{Accept the $a_{k+}$}
			\State {\bf break}	 \Comment{check the next action seq.}	
		\ElsIf{$\frac{1}{m}\sum_{q=1}^l c(b^q_{k:k+L};\phi, \delta) < 1 - \epsilon - \frac{m-l}{m}$} \Comment{Outer constraint is violated}
			\State {\bf break}	 \Comment{check the next action seq.}	
		\EndIf  
		\EndFor
		\EndFor
		\For {each $a_{k+}$ in $S$}
		\State expand missing laces and get $\hat{\mathcal{U}}^{(m)}(b_k, a_{k+})$
		\If {$\hat{\obj}^{*}_{(m)} < \hat{\mathcal{U}}^{(m)}(b_k, a_{k+})$}
		\State $a^*_{k+} \leftarrow a$, $\hat{\obj}^{*}_{(m)} \leftarrow 	\hat{\mathcal{U}}^{(m)}(b_k, a_{k+})$ 
		\EndIf
		\EndFor
		\State {\bf Return} $a^*_{k+}$
	\end{algorithmic}
\end{algorithm}

%% file: floats/AlgBruteForce.tex
\begin{algorithm}[t]
	\begin{algorithmic}[1]
		\caption{Optimality under probabilistic constraint (baseline) $\rho (\cdot) \equiv \phi(\cdot)$, $\mathcal{U}(\cdot) \equiv \mathcal{C}(\cdot)$}
		\label{alg:Baseline1}
		\State {\bf Input:} $\mathcal{A}$ 
		\State $a^*_{k+} \leftarrow \mathrm{undef}$, $\hat{\obj}^{*}_{(m)} \leftarrow -\infty$, 
		\For {each $a_{k+}$ in $\mathcal{A}$}
		\State Expand $m$ laces and get $\hat{\mathcal{U}}^{(m)}(b_k, a_{k+})$
		\If {$\hat{\obj}^{*}_{(m)} < \hat{\mathcal{U}}^{(m)}(b_k, a_{k+})$}
		\State $a^*_{k+} \leftarrow a_{k+}$, $\hat{\obj}^{*}_{(m)} \leftarrow 	\hat{\mathcal{U}}^{(m)}(b_k, a_{k+})$ 
		\EndIf
		\EndFor
		\If{$\hat{\obj}^{*}_{(m)} \geq \delta$ } 
		\State {\bf Return} $a^*_{k+}$
		\Else
		\State {\bf Return} No feasible action sequence present
		\EndIf
	\end{algorithmic}
	
\end{algorithm}

%% file: floats/Alg_purely_constr.tex
\begin{algorithm}[t]
	\begin{algorithmic}[1]
		\caption{Maximal feasible return (Bisection method)}
		\label{alg:PurelyConstrained}  
		\State {\bf Input:} $\mathcal{A}$, $\delta$ $\Delta$, $m$  
		\State $S \leftarrow \mathcal{A}$, $T \leftarrow \mathcal{A}$, $\tilde{\delta} \leftarrow (\delta + \Delta)\cdot 0.5$  
		\State $\forall  a_{k+} \in \mathcal{A} $ expand a single lace and $\tilde{m} \leftarrow 1$ \Comment{warm up}
		\While{true} \Comment{Trials loop}
		\For {{\bf each} $a_{k+} \in S$}
		\If {!\Call{AdaptBounds}{$a_{k+}$, $\tilde{m}$}}
		\State $S \leftarrow S \setminus a_{k+}$,
		\EndIf
		\EndFor
		\If{$|S| == 1$}
		\State \Return $a_{k+} \in S, \tilde{\delta}$
		\ElsIf{$S \subset \emptyset$}
		\State $S \leftarrow T$, $\Delta \leftarrow \tilde{\delta}$, $\tilde{\delta} \leftarrow (\delta +\tilde{\delta})\cdot 0.5$
		\State {\bf next}
		\Else 
		\State $T \leftarrow S$, $\delta \leftarrow \tilde{\delta}$, $\tilde{\delta} \leftarrow ( \tilde{\delta}+\Delta)\cdot 0.5$
		\EndIf 
		\EndWhile 
		\Procedure{AdaptBounds}{action seq: $a_{k+}$, counter: $\tilde{m}$}
		\While {  true }
		\If{$\tilde{m} < m$ }
		\State $\tilde{m} \leftarrow \tilde{m}+1$, Draw a lace $z^{\tilde{m}}_{k+1:k+L}$, 
		\EndIf 
		\If {$\frac{1}{m}\sum_{l=1}^{\tilde{m}} c(b^l_{k:k+L};\phi, \tilde{\delta}) < 1 - \epsilon - \frac{m-\tilde{m}}{m}$}  
		\State status $\leftarrow$ false
		\State {\bf break }
		\ElsIf{$1-\epsilon \leq \frac{1}{m}\sum_{l=1}^{\tilde{m}} c(b^l_{k:k+L};\phi, \tilde{\delta})$}
		\State status $\leftarrow$ true
		\State {\bf break }
		\EndIf
		\EndWhile
		\State \Return status
		\EndProcedure	
	\end{algorithmic}
\end{algorithm}

%% file: floats/Alg_baseline.tex
\begin{algorithm}[t]
	\begin{algorithmic}[1]
		\caption{Baseline maximizing $\widehat{\mathrm{VaR}}^{(m)}_{\epsilon}$}
		\label{alg:Baseline2}  
		\State {\bf Input:} $\mathcal{A}$
		\State $a^*_{k+} \leftarrow \mathrm{undef}$, $\hat{\obj}^{*}_{(m)} \leftarrow -\infty$
		\For {{\bf each} $a_{k+} \in \mathcal{A}$}
		\State {Expand $m$ laces and approximate $\widehat{\mathrm{VaR}}^{(m)}_{\epsilon}$}
		\If {$\hat{\obj}^{*}_{(m)} < \widehat{\mathrm{VaR}}^{(m)}_{\epsilon}$}
		\State $a^*_{k+} \leftarrow a_{k+}$, $\hat{\obj}^*_{(m)} \leftarrow 	\widehat{\mathrm{VaR}}^{(m)}_{\epsilon}$ 
		\EndIf
		\EndFor
		\State {\bf Return} $a^*_{k+}$ 
	\end{algorithmic}
\end{algorithm}

%% file: floats/tableOptimalityUnderProbConstraint.tex
\begin{table*}[t]
	\caption{Optimality under probabilistic constraint. Here we set $300$ observation laces per path. Each quantity was averaged over $5$ trials.}
	\centering
	\resizebox{\textwidth}{!}{
		\begin{tabular}{|c|c|c|c|c|c|c|c|c|c|c|c|}
			\hline
			&    $\epsilon$  & $\delta$ & $\mathcal{P}^{*}$& $\hat{\obj}^{*}_{(m)}$  & \textnumero discarded paths &time [sec] $\pm$ std& speedup \eqref{eq:speedup} &  laces frac. \eqref{eq:LacesFrac}     &total laces &\textnumero paths &\textnumero  land.     \\
			\hline
			Alg.~\ref{alg:Baseline1}  & - & $0.0$& $14$& $36.98\cdot 10^{-5}$& -& $1171.21 \pm 74.48$&   &  $0$ & $9000/9000$ &\multirow{7}{*}{$30$}   &\multirow{7}{*}{$4$} \\
			\cline{1-10}
			Alg.~\ref{alg:StochasticExploration}  & $0.023$& $0.0$ & no feasible& - & $30$& $77.67 \pm 4.01$ &$0.934$&$0.95$& $459/9000$& &\\ 
			\cline{1-10}   
			Alg.~\ref{alg:StochasticExploration}  & $0.3$& $0.0$ & $14$ &$36.98 \cdot 10^{-5}$& $29$& $489.44 \pm 26.46$ &$0.58$&$0.60$& $3559/9000$& &\\ 
			\cline{1-10}
			Alg.~\ref{alg:StochasticExploration}  & $0.5$&$0.0$& $14$ &$36.98\cdot 10^{-5}$& $29$ & $813.29 \pm 34.27$ &$0.31$& $0.37$ &$5685/9000$&& \\ 
			\cline{1-10}
			Alg.~\ref{alg:StochasticExploration}  & $0.7$& $0.0$& $14$& $36.98\cdot 10^{-5}$ & $27$&$974.18 \pm 51.14$& $0.17$ &$0.134$&$7794/9000$ && \\ 
			\cline{1-10}
			Alg.~\ref{alg:StochasticExploration}  & $0.8$& $0.0$& $14$& $36.98\cdot 10^{-5}$& $23$ & $1099.98 \pm 45.41$&$0.06$&$0.029$&$8738/9000$&& \\ 
			\cline{1-10}
			Alg.~\ref{alg:StochasticExploration}  & $0.9$& $0.0$& $14$& $36.98\cdot 10^{-5}$& $0$ & $1130.77 \pm 56.47$&$0.03$&$0.0$&$9000/9000$&& \\ 
			\cline{1-12}
	\end{tabular}
}
	\label{tbl:OptimalityUnderProbConstraint} 
\end{table*}

%% file: floats/table.tex
\begin{table*}[t]
	\caption{Solving maximum feasible return problem on top of $30$ candidate paths (Fig.~\ref{fig:paths}) with scenario presented in Fig.~\ref{fig:warmup2}. In this study the number of observation laces is $64$ per path. We observe that the speedup is approximately as the fraction of expanded laces, as expected since it is a little overhead from the adaptation. The values of time are averaged over $10$ trials. The speedup is calculated from mean planning time.}
	\centering
	\resizebox{\textwidth}{!}{
	\begin{tabular}{|c|c|c|c|c|c|c|c|c|c|c|}
		\hline
		 &    $\epsilon$  & $\mathcal{P}^{*}$& $\delta^{*}$  &time [sec] $\pm$ std & speedup \eqref{eq:speedup} &  laces frac. \eqref{eq:LacesFrac} &laces evaluations  &\textnumero paths with $\mathrm{VaR_{\epsilon}} > 0$  &\textnumero paths &\textnumero \ landmarks    \\
		 \hline
		 Alg.~\ref{alg:Baseline2}  & \multirow{2}{*}{$0.3$} & $4$& $9.71\cdot 10^{-6}$& $2568.97 \pm 6.32$& -&  $0$  & $1920/1920$ & \multirow{2}{*}{$18$} &\multirow{6}{*}{$30$}  & \multirow{6}{*}{$18$} \\
		 \cline{1-1} \cline{3-8}  
		  Alg.~\ref{alg:PurelyConstrained} && $4$ &$8.93\cdot 10^{-6}$&$1515.44\pm 27.35$&  $0.41$& $0.57$ & $830/1920$&&&\\
		 \cline{1-9} 
		  Alg.~\ref{alg:Baseline2}  & \multirow{2}{*}{$0.5$} & $5$ &$1.06\cdot 10^{-5}$&   $3233.26 \pm 65.59$ &  - & $$ & $1920/1920$ & \multirow{2}{*}{$19$} &&\\
		 \cline{1-1} \cline{3-8} 
		 Alg.~\ref{alg:PurelyConstrained} &&$5$ & $1.09\cdot 10^{-5}$& $2201.36 \pm 38.93$ & $0.32$ & $0.35$ &$1250/1920$&  &&\\
		 \cline{1-9} 
		 Alg.~\ref{alg:Baseline2}  & \multirow{2}{*}{$0.7$} & $5$ &$1.73\cdot 10^{-5}$&$3299.26 \pm 71.26$ & &  $$    &$1920/1920$& \multirow{2}{*}{$20$}& &\\
		 \cline{1-1} \cline{3-8}
		 Alg.~\ref{alg:PurelyConstrained} && $5$ & $1.59\cdot 10^{-5}$&$2433.06 \pm 46.19$ & $0.26$ &$0.27$&$1411/1920$ &&&\\
		 \cline{1-11}    
	\end{tabular}
}
	\label{tbl:MaximalFeasibleReturn} 
\end{table*}

%% file: floats/table_statistical.tex
\begin{table*}[t]
	\caption{Analysis of the behavior with randomly drawn landmarks. Each trial we randomly drawn $10$ landmarks in the square $[2, 5] \times [2, 5]$. Here the visibility radius of the landmarks is $0.8$. Note that mean time based speedup and the accumulated time based speedup are identical since the difference in running time in two possibilities is only the division by the number of trials. }
	\centering
	\begin{tabular}{|c|c|c|c|c|c|}
		  \hline
		   \textnumero \ paths & $30$ & $30$ & $30$ & $30$ & $30$  \\ 
		  \hline 
		  accumulated time based speedup & $0.35$ &$ 0.26$ &$0.21$&$0.14 $& $0.055$ \\
		  \hline 
		  min speedup &$0.14$ &$0.092$ &$0.14$&$0.08$&$0.019$\\
		  \hline
		  max speedup &$0.57$ &$0.43$ &$0.32$&$0.22$&$0.081$\\
		  \hline
		  mean time based speedup & $0.35$ &$0.26$ &$0.21$ & $0.14$ & $0.055$ \\ 
		  \hline
		  mean time [sec] $\pm$ std Alg.~\ref{alg:Baseline2} & $817.79 \pm 132.12$&$771.59 \pm 119.82$ &$1670.69 \pm 260.39$  & $1607.60 \pm 248.46$& $1671.69 \pm 259.52$\\
		  \hline
		  mean time [sec] $\pm$ std Alg.~\ref{alg:PurelyConstrained} & $530.47 \pm 162.30$ &$574.55 \pm 130.11$ &$1320.71 \pm  216.86$ &$1382.89 \pm 245.19$ &$1580.00 \pm 230.63$ \\
		  \hline
		  accumulated time [sec] Alg.~\ref{alg:Baseline2}  & $8177.92$ &$7715.88$&$16706.90$&$16075.98$& $16716.86$\\
		  \hline
		  accumulated time [sec]  Alg.~\ref{alg:PurelyConstrained} &$5304.69$ &$5745.51$ &$13207.11$&$13828.99$& $15800.04$\\ 
		   \hline
		  accumulated skipped laces frac. &$0.36$ &$0.29$&$0.23$&$0.15$ &$0.065$\\
		  \hline
		  accumulated expanded laces Alg.~\ref{alg:PurelyConstrained}&$12285$ &$13689$&$14800$&$16304$& $17944$\\ 
		  \hline
		  total \textnumero \ of laces & $19200$ &$19200$ &$19200$ & $19200$&$19200$\\
		  \hline
		  \textnumero trials &$10$ & $10$ & $10$& $10$ &$10$\\ 
		 \hline 
		 \textnumero \ landmarks & $10$ & $10$ &$10$ &$10$ & $10$\\
		 \hline
		 $\epsilon$ & 0.2 &$0.3$ & $0.5$ & $0.7$& $0.9$\\
		 \hline 
		 $\delta^{\text{min}}$& $0.0$ & $0.0$ &$0.0$&$0.0$& $0.0$\\
		 \hline
	\end{tabular}
	\label{tbl:statistical} 
\end{table*}